\pdfoutput=1

\documentclass[11pt]{article}
\usepackage{graphicx}
\usepackage{EMNLP2023}

\usepackage{tabularx}
\usepackage{amsmath}
\usepackage{amsthm}
\usepackage{xcolor}
\usepackage{colortbl}
\usepackage{svg}
\usepackage{amssymb}
\usepackage{times}
\usepackage{latexsym}
\usepackage{multirow}
\usepackage[normalem]{ulem}
\useunder{\uline}{\ul}{}
\usepackage{array}
\usepackage{bbm}
\usepackage{bm}
\usepackage{algorithm}
\usepackage{booktabs}
\usepackage{siunitx}
\usepackage{multicol}
\usepackage[noend]{algpseudocode}

\usepackage[T1]{fontenc}

\usepackage[utf8]{inputenc}

\usepackage{microtype}
\usepackage{listings}
\usepackage{inconsolata}
\usepackage{subcaption}

\title{Weights-Rotated Preference Optimization for Large Language Models}

\author{Chenxu Yang\textsuperscript{\rm 1,2}\thanks{\ \ \ Equal contribution. }, Ruipeng Jia\textsuperscript{\rm 3}\footnotemark[1], Mingyu Zheng\textsuperscript{\rm 1,2}, Naibin Gu\textsuperscript{\rm 1,2}, Zheng Lin\textsuperscript{\rm 1,2}\thanks{\ \ \ Zheng Lin is the corresponding author. },  \\
{\bf Siyuan Chen\textsuperscript{\rm 1,2}, Weichong Yin\textsuperscript{\rm 3}, Hua Wu\textsuperscript{\rm 3}, Weiping Wang\textsuperscript{\rm 1} }  \\
  \textsuperscript{\rm 1}Institute of Information Engineering, Chinese Academy of Sciences, Beijing, China \\
  \textsuperscript{\rm 2}School of Cyber Security, University of Chinese Academy of Sciences, Beijing, China \\
  \textsuperscript{\rm 3}Baidu Inc., Beijing, China \\
  \texttt{\textrm{\{}yangchenxu,linzheng,wangweiping\textrm{\}}@iie.ac.cn} \\
  \\
  }

\begin{document}
\maketitle

\begin{abstract}

Despite the efficacy of Direct Preference Optimization (DPO) in aligning Large Language Models (LLMs), reward hacking remains a pivotal challenge. This issue emerges when LLMs excessively reduce the probability of rejected completions to achieve high rewards, without genuinely meeting their intended goals. As a result, this leads to overly lengthy generation lacking diversity, as well as catastrophic forgetting of knowledge. We investigate the underlying reason behind this issue, which is representation redundancy caused by neuron collapse in the parameter space. Hence, we propose a novel Weights-\textbf{Ro}tated \textbf{P}reference \textbf{O}ptimization (\textbf{RoPO}) algorithm, which implicitly constrains the output layer logits with the KL divergence inherited from DPO and explicitly constrains the intermediate hidden states by fine-tuning on a multi-granularity orthogonal matrix. This design prevents the policy model from deviating too far from the reference model, thereby retaining the knowledge and expressive capabilities acquired during pre-training and SFT stages. Our RoPO achieves up to a 0.5-point improvement on AlpacaEval 2, and surpasses the best baseline by 1.9 to 4.0 points on MT-Bench with merely 0.015\% of the trainable parameters, demonstrating its effectiveness in alleviating the reward hacking problem of DPO.

\end{abstract}

\section{Introduction}

\begin{figure}[!t]
  \centerline{\includegraphics[scale=0.23]{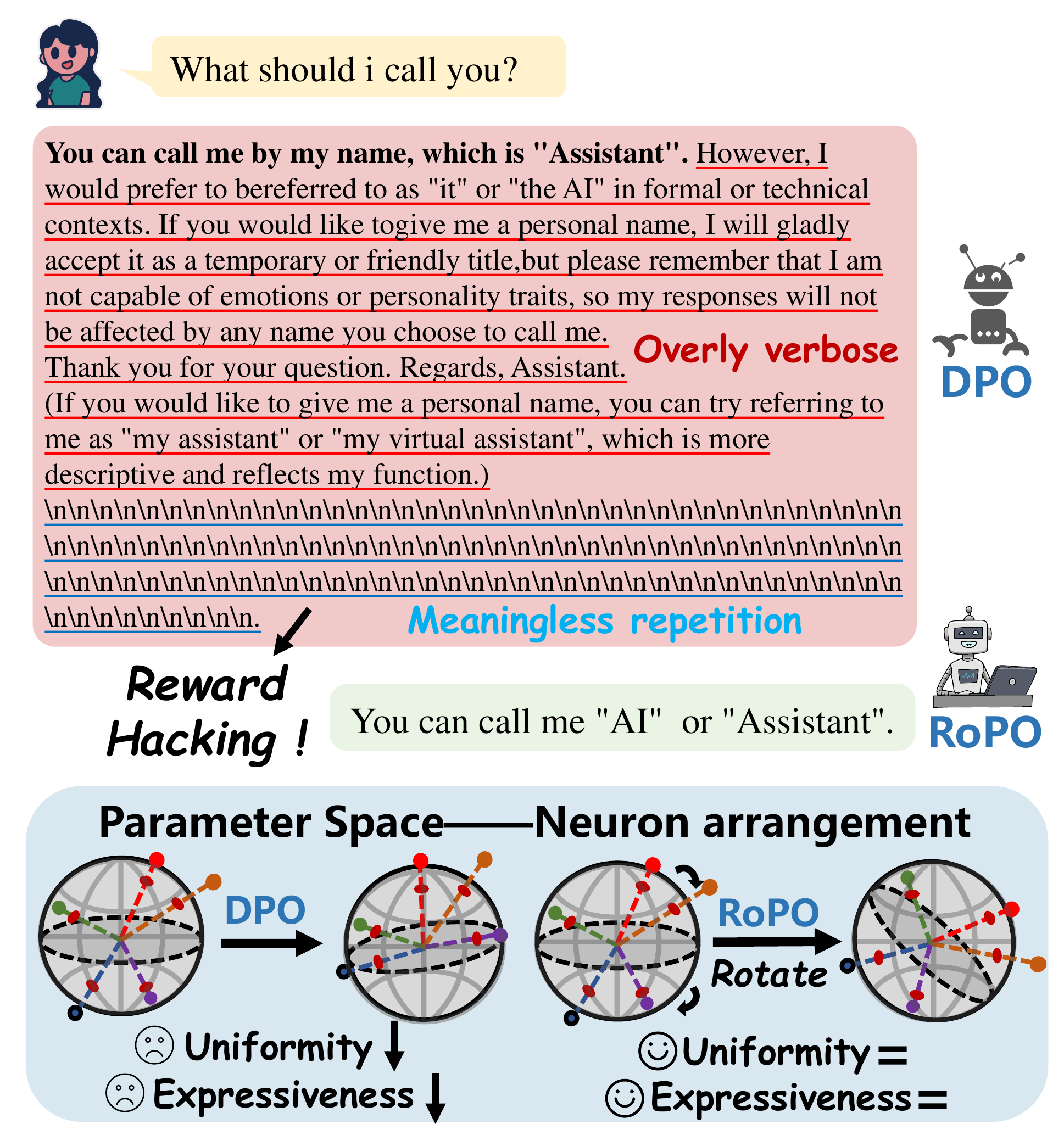}}
  \caption{DPO suffers from reward hacking, causing the model's expressive capability to decline and generating overly long content with meaningless repetitions. The isotropic distribution of the neurons was disrupted during DPO. RoPO mitigates it by rotary tuning, which retains the angle-encoded expressive knowledge.}
  \label{fig-motivation}
\end{figure} 
Despite achieving remarkable performance, large language models (LLMs) \cite{chatgpt,touvron2023llama2,bai2023qwen,yang2023baichuan,yang2025testtimepromptintervention,yang2025dynamicearlyexitreasoning} still pose the risk of generating content that diverges from human expectations \cite{bai2022constitutionalaiharmlessnessai}. To tackle this challenge, researchers introduced reinforcement learning from human feedback (RLHF) to improve the controllability of AI systems by simulating human preferences across various response options \cite{christiano2023deepreinforcementlearninghuman,ouyang2022traininglanguagemodelsfollow,stiennon2022learningsummarizehumanfeedback,dai2025sgrpoearlyexitreinforcement}. However, RLHF has been widely criticized for its training instability and high sensitivity to hyperparameters. Recently, researchers have developed RL-free direct alignment algorithms \cite{dong2023raftrewardrankedfinetuning,yuan2023rrhfrankresponsesalign,zhao2023slichfsequencelikelihoodcalibration}, with Direct Preference Optimization (DPO) standing out as a leading approach in this field.

 

Unfortunately, there exists a significant drawback in DPO: \textit{reward hacking}, also known as \textit{reward overoptimization}. Deviating from the primary goal of learning the characteristics from preferred completions and discouraging undesirable behaviors in rejected ones, the policy model exploits vulnerabilities in the DPO function by overly suppressing the likelihood of rejected completions to maximize rewards. Consequently, the fine-tuned model suffers from uncontrollable length, diminished diversity, and knowledge forgetting \cite{shen-etal-2023-loose,RDPOpark2024disentanglinglengthqualitydirect}. As shown in Figure \ref{fig-motivation}, the DPO-trained model produces overly verbose output with meaningless repetitions when answering simple queries. Recently, several approaches have been introduced to address this challenge by modifying the constraints in the loss function \cite{azar2023generaltheoreticalparadigmunderstand,wang2023reverseklgeneralizingdirect,zeng2024tokenleveldirectpreferenceoptimization,dpop,ORPOhong2024orpomonolithicpreferenceoptimization}. However, these methods typically degrade alignment performance and perform worse than DPO.




In this paper, we investigate the underlying reasons behind the reward overoptimization of DPO from the perspective of parameter space. Our findings in Section \ref{pilot-exp} indicate that, compared to supervised fine-tuned (SFT) training, DPO optimization causes the model's neurons to collapse in the parameter space, leading to representation redundancy issues. To overcome this problem, we propose {W}eights-\textbf{Ro}tated \textbf{P}reference \textbf{O}ptimization (\textbf{RoPO}) algorithm, which simultaneously imposes constraints on both intermediate hidden layers from the parameter perspective and output layer from the logits perspective. The output layer is implicitly constrained using the original KL divergence from DPO to maintain the diverse and fluent expressions of the SFT model semantically, while the intermediate hidden layers are explicitly constrained by rotary-tuning with a multi-granularity orthogonal matrix to preserve angle-encoded knowledge. The dual constraints together prevent the policy model from deviating too far from the reference model, retaining the knowledge and expressions acquired during pre-training and SFT stages. Specifically, the specially designed multi-granularity orthogonal matrix consists of fine-grained Givens matrices and global Householder Reflection matrices.

Extensive experiments show that RoPO achieves strong alignment performance while preventing excessively long generations, repetitive expressions, and catastrophic forgetting of knowledge. RoPO consistently outperforms all preference optimization baselines in all benchmarks with merely 0.015\% of the trainable parameters. Specifically, RoPO surpasses the best baseline by 1.9 to 4.0 points on MT-Bench. By analyzing the training process, we discovered that the complementary constraints of RoPO prevent excessive reduction of the rejected completions likelihood while achieving higher reward accuracy, effectively mitigating the reward hacking problem.


Our contributions are summarized as follows:
 \begin{itemize}
	\item We delve into the underlying causes behind DPO reward hacking from a novel perspective.
	\item We propose an innovative method RoPO, which simultaneously imposes logit regularization and weight regularization to alleviate the reward hacking problem in DPO. 
	\item The extensive experimental results across five benchmarks validate that our RoPO is comprehensively effective, with significantly reduced training parameters and training efficiency. 
\end{itemize}





\section{Related Work}
\paragraph{Direct Preference Optimization.}

With the widespread application of LLMs, aligning them with human preferences has gained significant attention. Due to training instability and hyperparameter sensitivity in RLHF, recent studies have introduced several RL-free preference optimization methods \cite{dong2023raftrewardrankedfinetuning, zhao2023slichfsequencelikelihoodcalibration, yuan2023rrhfrankresponsesalign}. \citet{rafailov2023directpreferenceoptimizationlanguage} derived Direct Policy Optimization (DPO) theoretically by fitting an implicit reward function via reparameterization. As the most influential and effective approach, DPO significantly lowers the alignment barrier for LLMs.
Numerous studies followed the proposition of DPO: IPO \cite{azar2023generaltheoreticalparadigmunderstand} revised the loss to minimize the disparity between the ratio of log-likelihoods and a given threshold. KTO \cite{KTOethayarajh2024ktomodelalignmentprospect} directly maximized the utility of generations. \citet{meng2024simposimplepreferenceoptimization} proposed SimPO, considering the average log-probability of a sequence as the implicit reward. Others aim to resolve different issues within the DPO objective function \cite{ORPOhong2024orpomonolithicpreferenceoptimization, CPOxu2024contrastivepreferenceoptimizationpushing, qi2024onlinedpoonlinedirect, wang2023reverseklgeneralizingdirect, RDPOpark2024disentanglinglengthqualitydirect,RSOliu2024statisticalrejectionsamplingimproves, zhou2024wpoenhancingrlhfweighted}. Among these, the four studies most relevant to ours are DPOP \cite{dpop}, SamPO \cite{lu2024eliminatingbiasedlengthreliancesampo}, LD-DPO \cite{liu2024lengthdesensitizationdirectpreferencelddpo} and InterDPO \cite{kojima2024intermediatedirectpreferenceoptimization}. DPOP, SamPO, LD-DPO also focused on the reward hacking problem in DPO and attempted to fix the failure modes by adding a penalty term; InterDPO concentrated on the optimization of intermediate layers by calculating DPO loss at these layers. Different from their method, we incorporate explicit regularization at the intermediate layers from parameter perspective.



\paragraph{Orthogonal Regularization.} Orthogonal regularization has emerged as a powerful technique in deep learning, which is prevalent in research across the NLP \cite{Mao_2020, smith2017offlinebilingualwordvectors, conneau2018wordtranslationparalleldata} and CV \cite{Brock2016NeuralPE, Brock2018LargeSG} fields. It is widely used to alleviate gradient vanishing and explosion by maintaining a constant norm \cite{Brock2018LargeSG} or to preserve the geometric structure of word vectors, thereby retaining semantic information within them \cite{smith2017offlinebilingualwordvectors}. In contrast to their approaches, our RoPO introduces an orthogonal matrix from a parameter perspective for regularization rather than loss regularization, avoiding excessive deviation of the policy model during DPO training.

\section{Preliminaries}


\subsection{Hyperspherical Energy}
Hyperspherical energy (HE) was originally proposed to measure the uniformity of neuron arrangement in high-dimensional space \cite{liu2020learningminimumhypersphericalenergy}. Higher HE indicates that neurons collapse in closely related directions, leading to representation redundancy; while lower HE suggests a more uniform arrangement, indicating better representation ability of the model. Suppose that there is a fully connected layer $\boldsymbol{W=}\{\boldsymbol{w}_1,\cdots,\boldsymbol{w}_n\}\boldsymbol{\in}\mathbb{R}^{d\times n}$, where $\boldsymbol{w}_i\in\mathbb{R}^d$ denotes the $i$-th neuron. The definition of HE is as follows:
\begin{equation}
{\text{HE}}(\boldsymbol{W}) = \sum_{i\neq j}\|\hat{\boldsymbol{w}}_i-\hat{\boldsymbol{w}}_j\|^{-1},
\label{he}
\end{equation}
where $\boldsymbol{\hat{w}}_i=\frac{\boldsymbol{w}_i}{\|\boldsymbol{w}_i\|}$ is the $i$-th normalized neuron.

\begin{figure}[!t]
  \centerline{\includegraphics[scale=0.33]{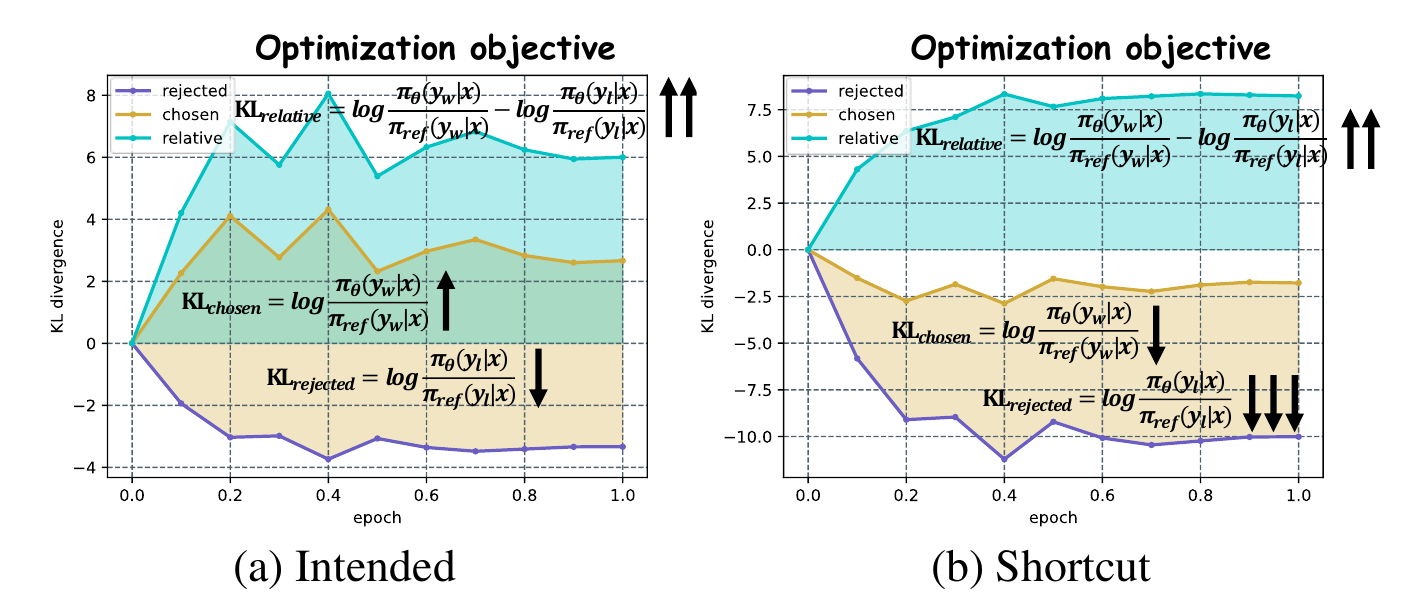}}
  \caption{The training objective was achieved through more excessive suppression on the rejected completions than on the chosen completions in DPO, suggesting a potential issue of reward hacking. }
  \label{pilot-kl}
\end{figure} 

\subsection{DPO and its Reward Hacking}
Direct Preference Optimization (DPO) is an optimization method that directly learns the policy bypassing the reward function \cite{rafailov2023directpreferenceoptimizationlanguage}.
\begin{equation}
\scalebox{0.85}{$
\begin{aligned}
&\mathcal{L}_{\mathrm{DPO}}(\pi_\theta;\pi_{\mathrm{ref}})=\\
&-\mathbb{E}\left[\log\sigma\left(\beta\log\frac{\pi_\theta(y_w\mid x)}{\pi_{\mathrm{ref}}(y_w\mid x)}-\beta\log\frac{\pi_\theta(y_l\mid x)}{\pi_{\mathrm{ref}}(y_l\mid x)}\right)\right],
\end{aligned}$}
\end{equation}
where $x$ denotes the prompt, $y_w$ denotes the chosen completion, $y_l$ denotes the rejected completion, $\pi_\theta$ is the policy model, and $\pi_{\mathrm{ref}}$ is the reference model.

The objective of the DPO function could be regarded as optimizing the relative probability of choosing the chosen completion over the rejected one $p^*(y_w\succ y_l|x)$ in each pair of samples.
\begin{equation}
p^*(y_w\succ y_l|x)= \sigma\left(\beta\log\frac{\pi_\theta(y_w\mid x)}{\pi_\theta(y_l\mid x)}-\gamma\right),
\end{equation}
where $\gamma=\beta\log\frac{\pi_{\mathrm{ref}}(y_w\mid x)}{\pi_{\mathrm{ref}}(y_l\mid x)}$ could be regarded as a constant as $\pi_{\mathrm{ref}}$ is not updated during training.


To minimize the DPO loss function, the policy model would try to increase the probability $p^*(y_w\succ y_l|x)$, that is, increasing the ratio $\frac{\pi_\theta(y_w\mid x)}{\pi_\theta(y_l\mid x)}$. The original intention of human preference alignment was to learn the features that humans prefer in the chosen completions and suppress the undesirable behaviors in the rejected completions. However, subject to an upper bound 1, it is difficult to significantly modify $\pi_\theta(y_w|x)$ to increase the probability ratio. As a result, the policy model turns to push the probability of $y_l$ as low as possible to achieve high rewards, leading to overoptimizing the spurious pattern of suppressing rejected examples. In some cases, DPO could lead to a reduction in the likelihood of chosen completions when the pair of completions holds small edit distances \cite{dpop}. We verified this by observing KL divergences alterations over 1 epoch in DPO-tuning on a held-out set. Figure \ref{pilot-kl} illustrates that the relative likelihood increases through more excessive suppression on the rejected completions than on the chosen completions.  Reward hacking suppresses all behavioral characteristics in rejected completions, resulting in poor expressive ability and the deterioration of generation diversity. 




\section{Methodology}

The Kullback-Leibler (KL) divergency in DPO prevents the policy model from deviating too far from the arrangement on which the reward model is accurate, as well as maintaining generation diversity and preventing mode-collapse to single high-reward answers \cite{rafailov2023directpreferenceoptimizationlanguage}.  However, they merely regulate updates on the output logits implicitly. We aim to add an explicit constraint from the perspective of parameter updates to suppress deviations, combining it with the implicit KL divergency to prevent reward hacking, enabling LLMs to better align human preference without sacrificing expressiveness power.

\subsection{Parameter Perspective Analysis} \label{pilot-exp}

To better design this explicit regularization, we conducted experiments to explore how DPO's parameter updates affect the model's expressiveness. 
As \citet{liu2020learningminimumhypersphericalenergy} pointed out, the degree of uniformity in neuron placement influences model's expressiveness, and representation redundancy harms model performance. \citet{qiu2024controllingtexttoimagediffusionorthogonal} held the view that some of the model's knowledge is contained within the relative angles between neurons. Drawing inspiration from their research, we hypothesize that the reward hacking problem in DPO is caused by the collapse of neuron arrangement, resulting in knowledge forgetting and verbose generation. 

\begin{figure}[!t]
    \centering
    \begin{minipage}{0.25\textwidth}
        \centering
        \includegraphics[width=\linewidth]{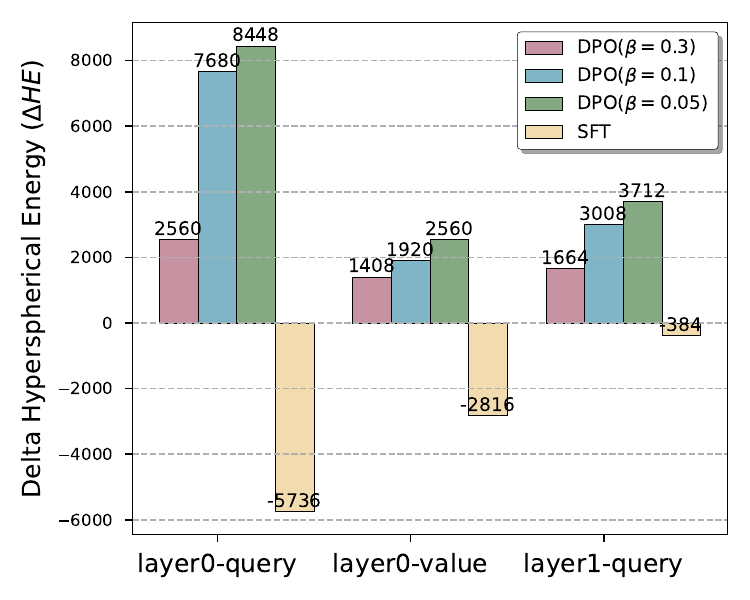}
        \subcaption{Hyperspherical Energy variations. Increased HE indicates neuron arrangement collapse.}
        \label{sub-he}
    \end{minipage}\hfill
    \begin{minipage}{0.195\textwidth}
        \centering
        \includegraphics[width=\linewidth]{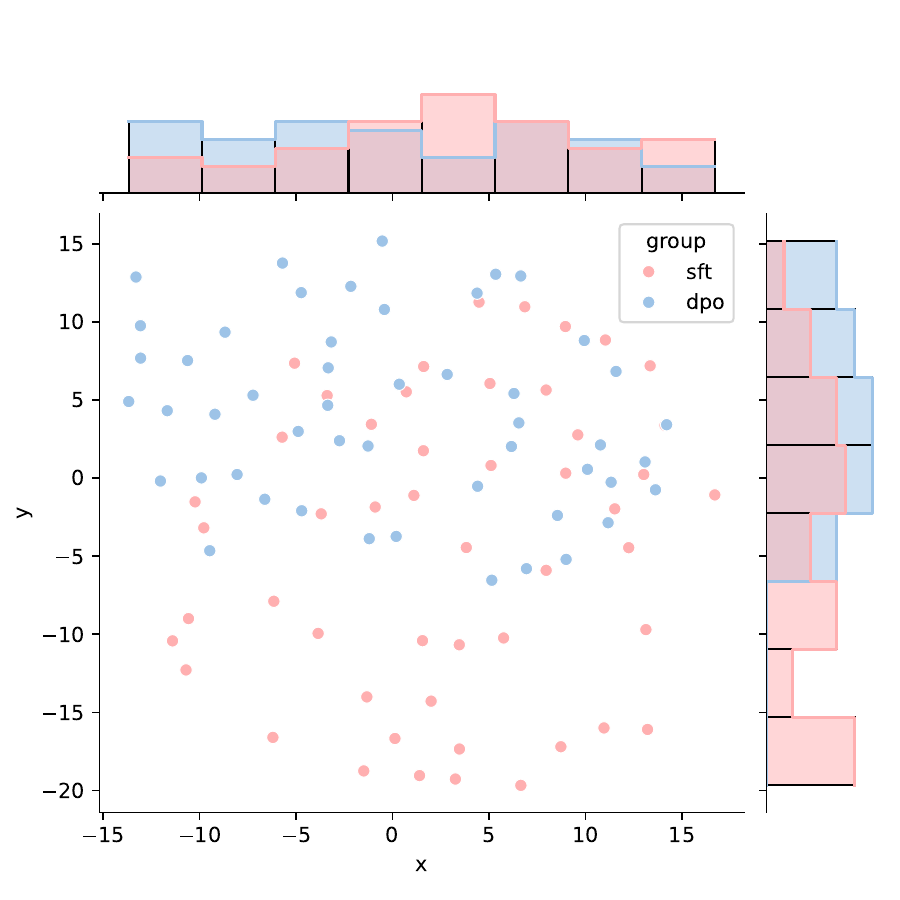}
        \subcaption{Neuron arrangement changes of query vector in layer 0 after DPO.}
        \label{sub-tsne}
    \end{minipage}

    \caption{Hyperspherical Energy variations of the query and value vectors in layer 0 and layer 1 after training.}
    \label{figure-he}
\end{figure}

To validate our hypothesis, we compared the changes in hyperspherical energy of some layers in models trained by DPO and SFT, and used t-SNE visualization to observe the differences in neuron arrangement before and after DPO \cite{JMLR:v9:vandermaaten08a}. Figure \ref{sub-he} shows that SFT reduces the model's hypersphere energy, while DPO leads to an increase. This suggests that the model learns new knowledge during the SFT stage, resulting in a more uniform arrangement of neurons. The reward hacking of DPO impairs expressive knowledge encoded in neuron angles.
Moreover, with the weakening of the KL constraint ($\beta\downarrow$), the increase in hyperspherical energy becomes more notable, indicating that DPO's overoptimization is positively correlated with representation redundancy. From the visualization presented in Figure \ref{sub-tsne}, we observe that the neurons are more densely arranged along the y-axis after DPO, which is consistent with the result of hyperspherical energy. 

\captionsetup[subfigure]{labelformat=parens}

\subsection{Weights-Rotated Preference Optimization}

Based on the above observations, we propose our Weights-\textbf{Ro}tated \textbf{P}reference \textbf{O}ptimization (\textbf{RoPO})  by adding an orthogonal regularization on intermediate hidden layers from the parameter perspective, while keeping the KL divergence regularization on output layer. The dual constraints together prevent the policy model from deviating too far from the reference model in the alignment process, retaining the angle-encoded knowledge and expressions acquired during pre-training and SFT stages. In the following, we introduce detailed design of the regularization on intermediate hidden layers.

\begin{figure*}[htbp]
  \centerline{\includegraphics[scale=0.38]{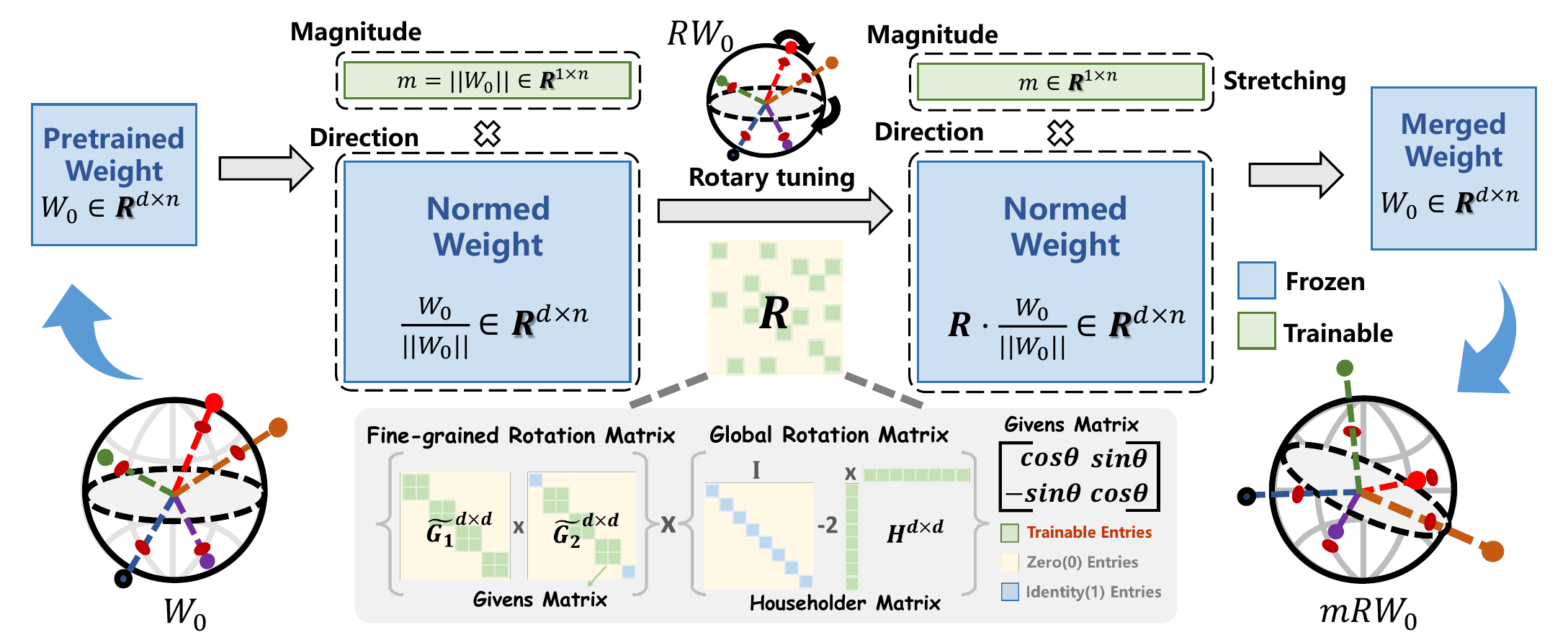}}
  \caption{An overview of our RoPO method. }
  \label{fig1}
\end{figure*}

\paragraph{Weight Decomposition}

Denoting $\bm{W}=\{\bm{w_1},\dots,\bm{w_n} \}\in \mathbb{R}^{d\times n }$ as the weight of policy model, we regard $\bm{w_i}\in \mathbb{R}^{d}$ as the $i$-th neuron. RoPO decomposes the weight into its magnitude vector $\bm{m}=\{m_1,\dots,m_n \}\in \mathbb{R}^{1\times n }$ and directional matrix $\bm{W}/\|  \bm{W} \|_c$, where $\| \cdot \|_c$ is the vector-wise norm of a matrix across each column, ensuring each neuron in the directional matrix remains a unit vector. We initialize the magnitude vector $\bm{m}$ as $\|  \bm{W} \|_c$ and the policy model' weight $\bm{W}$ with the SFT model' weight $\bm{W}_\text{SFT}$ to ensure continuity for preference optimization. 
\begin{equation}
\begin{aligned}
\bm{W}'=\bm{m}\cdot\frac{   \bm{W} }{\| \bm{W} \|_c}={\| \bm{W} \|_c}\cdot\frac{   \bm{W} }{\| \bm{W} \|_c}. 
\end{aligned}
\end{equation}

After decomposition, RoPO introduces a specially designed orthogonal matrix $\bm{R}\in \mathbb{R}^{d\times d }$ to tune the frozen directional weight $\bm{W}$ through rotation, keeping magnitude vector $\bm{m}$ trainable. Denoting $\bm{x} \in \mathbb{R}^d$ and $\bm{z} \in \mathbb{R}^n$ as the input and output vectors, the forward pass of RoPO is as follows:
\begin{equation}
\begin{aligned}
\bm{z}=\bm{W}'^\top \bm{x}=\underline{\bm{m}}\cdot(\underline{\bm{R}}\cdot\frac{ \bm{W} }{\| \bm{W} \|_c})^\top \bm{x}, \\  
\mathrm{s.t.} \,\bm{R}^\top\bm{R}=\bm{R}\bm{R}^\top=\bm{I},
\end{aligned}
\label{eq8}
\end{equation}
where $\bm{I}$ denotes an identity matrix, and the trainable parameters are denoted by an \underline{underline}.

As illustrated in Figure \ref{fig1}, RoPO regulates the weights updating by performing rotation across $d$ input dimensions and stretching each neuron $\bm{w_i}$ in the weight matrix to keep relative angular distances between weight vectors invariant between policy model and SFT model. The regularization ensures the angle-encoded knowledge acquired in the previous training stage preserves, which helps alleviate reward overoptimization. \footnote{RoPO does not introduce additional overhead during the inference stage since the parameter matrices can be merged.}  

\paragraph{Multi-Granularity Orthogonal Matrix} Specifically, we construct $\bm{R}$ with a combination of global rotation matrix $\tilde{\bm{H}} \in \mathbb{R}^{d\times d }$ and fine-grained rotation matrices $\tilde{\bm{G_1}}\tilde{\bm{G_2}}\in \mathbb{R}^{d\times d}$ as $R=\tilde{\bm{G_1}}\tilde{\bm{G_2}}\tilde{\bm{H}}$. We term it as multi-granularity orthogonal matrix, as the model weight is firstly rotated globally, and then rotated in each $2$-$d$ subspace planes locally. The orthogonality of the Multi-Granularity Orthogonal Matrix is formally proved in Appendix \ref{Ortho}.



 The global rotation matrix is composed by two Householder Reflection matrices, rotating the weight matrix globally with the following matrix:
\begin{equation}
\begin{aligned}
\tilde{\bm{H}}&=(\bm{I}-2\underline{\bm{u_1}}^\top\cdot \underline{\bm{u_1}})\cdot(\bm{I}-\underline{\bm{u_2}}^\top\cdot \underline{\bm{u_2}}),
\end{aligned}
\end{equation}
where $\bm{u_1}$ and $\bm{u_2}\in \mathbb{R}^{ d}$ are two trainable unit vectors initialized as $[1, 0,\dots,0]_{1\times d}$.

The fine-grained rotation matrices are composed by $d-1$ Givens rotation matrices $\bm{G}$ with arrangement as follows:
\begin{equation}
\begin{aligned}
\tilde{\bm{G_1}}&=\prod_{k=0}^{(d/2)-1}\bm{G}(2k,2k+1;\theta_k), \\
\tilde{\bm{G_2}}&=\prod_{k=0}^{(d-1)/2}\bm{G}(2k+1,2k+2;\theta_k),
\end{aligned}
\end{equation}
where the Givens matrix $\bm{G}$ rotates a vector in a 2-dimensional subspace planes, and the rotation angle is controlled by $\theta$. Suppose we have the following Givens matrix, where $\cos\theta$ appears at $\{(i, i), (j, j)\}$, $\sin\theta$ appears at $\{(i, j), (j, i)\}$. 
\begin{equation}
\bm{G}(i,j,\theta)=\begin{pmatrix}I&0&0&0&0\\0&\cos\theta&0&\sin\theta&0\\0&0&I&0&0\\0&-\sin\theta&0&\cos\theta&0\\0&0&0&0&I\end{pmatrix},
\end{equation}
where the trainable parameter $\theta $ is initialized as $0$ to ensure that $\bm{W}'$ equals to $\bm{W}$ before the finetuning. 


The arrangement of Givens matrices in $\tilde{\bm{G_1}}\tilde{\bm{G_2}}$ ensures any $d$-dimensional rotation could be accomplished, demonstrating its sufficient fitting capacity. In other words, given any vector $\boldsymbol{v}\in\mathbb{R}^d $, $\tilde{\bm{G_1}}\tilde{\bm{G_2}}$ could rotate it to any vector $\boldsymbol{y}\in\mathbb{R}^d$ on the same sphere with $\boldsymbol{v}$, providing a full-angle coverage. The proof of this statement is detailed in the Appendix \ref{frm}. Besides, the fine-grained rotation matrices could incorporate a rapid implementation approach of sparse matrix multiplication like in Rotary Position Embedding (RoPE) \cite{su2023roformerenhancedtransformerrotary}, accelerating the training process. The enhancement scheme is detailed in the Appendix \ref{accele}.




\begin{table*}[]
\centering
\renewcommand\arraystretch{0.93}
\setlength{\tabcolsep}{6pt}
\scalebox{0.8}{
\begin{tabular}{lccccccccccccccc}
\toprule
\multirow{3}{*}{\textbf{Method}} & \multicolumn{7}{c}{\textbf{Mistral-Base (7B)}}                                                                              & \multicolumn{7}{c}{\textbf{Llama2-Base (13B)}}                                                                               \\ \cline{2-15} 
                                 & \multicolumn{3}{c}{\textbf{AlpacaEval 2}} & \multicolumn{2}{c}{\textbf{Arena-Hard}} & \multicolumn{2}{c}{\textbf{MT-Bench}} & \multicolumn{3}{c}{\textbf{AlpacaEval 2}} & \multicolumn{2}{c}{\textbf{Arena-Hard}} & \multicolumn{2}{c}{\textbf{MT-Bench}} \\ \cline{2-15} 
                                 & \textbf{LC}  & \textbf{LWR}     & \textbf{Len.}      & \textbf{LWR}      & \textbf{Len.}     & \textbf{LWR}     & \textbf{Len.}   & \textbf{LC}   & \textbf{LWR}       & \textbf{Len.}      & \textbf{LWR}      & \textbf{Len.}     & \textbf{LWR}     & \textbf{Len.}    \\ \hline
                                 \rowcolor[rgb]{0.957,0.957,0.957}
SFT                           & 2.5   & 4.2                 & 790                & 5.4                & 1157              & 12.8              & 808       
&  2.3     & 3.7                 & 1093                & 4.4                & 1602             & 8.7              & 985             \\ 
DPO            & 9.7    & 10.2                & 1347               & 14.7               & 1614              & 14.8              & 1591        
&  8.0   & 9.0                & 1379              & 11.2             & 1949              & 15.0              & 1211             \\
SimPO         &   10.6        
& 10.0                & 1562  
& 13.7               & 1789             
& 17.3              & 1613  
& 8.4
& 9.3              & 1510         
& 13.5            & 2053         
& 12.5            & 1053             \\

R-DPO                 &  8.2          & 9.9                 & 1273               & 12.6               & 1642              & 9.9               & 1340      
&   7.9   & 9.3               & 1359               & 11.4               & 1894              & 7.3              & 1164           \\
KTO                    &    8.4       & 8.9                 & 1286               & 12.9               & 1475              & 14.7              & 1106     
&  4.8    & 5.2                 & 1529             & 13.1              & 2026             & 4.3              & 1295             \\
WPO                    &  10.7        &    11.5              & 1506               &    15.0           &   1763           &       20.3        &    1679    
&   8.0   &    9.4              &     1487         &     13.6          &       2029        &     17.8      & 1207           \\
DPO*          &   8.1    & 9.3                 & 1194               & 10.9               & 1470              & 13.8              & 1390       
&  4.7   & 6.4                & 1263               & 7.9               & 1809              & 11.5              & 1096             \\
IPO                    &      5.0     & 8.8                 & 1007               & 13.0               & 1347              & 11.1              & 929    
&    4.3     & 6.8               & 1313               & 9.7             & 1863              & 10.1              & 1186             \\
ORPO                    &     4.5     & 5.7                 & 1074               & 9.8                & 1345              & 13.4              & 992   
&    3.1   & 3.5                 & 1358               & 8.5               & 1824             & 10.6              & 1143            \\
DPOP           &     8.4     
& {9.2}                & 1238     
& {14.5}               & 1512     
& {16.8}              & 1501    
& 7.2
& 8.1               & 1355 
& 9.7     & 1826   
& {12.8}              & 1176            \\
Inter-DPO          &    8.0       
& 8.7                & 1429            
& 10.4               & 1611              
& 12.4              & 1731   
&     6.9      
& 6.6               & 1319               
& 9.6             & 2024              
& 11.2              & 1209              \\
DPO-LoRA            &  6.2       & 10.0                & 962                & 11.2               & 1421              & 16.0              & 920     
&  5.1      & 8.9                & 1267               & 9.7               & 1888              & 6.2              & 1133              \\
SamPO            &    11.2     &   13.0           &     1156       &      15.1        &  1583         &      22.1      &   1286  
&    7.4   &       9.2       &     1308        &   13.2        &       1848      &     14.2       &      1208      \\
LD-DPO            &    10.8     &   13.2         &      979       &        14.2      &   1401          &      20.6     &  1087  
&    7.3   &       9.7         &     1199       &    11.9    &       1823     &     18.4     &     1161
\\
\rowcolor[rgb]{0.87,0.94,1}
RoPO            &     \textbf{11.4}        & \textbf{13.5}                & 1042               & \textbf{16.0}               & 1433              & \textbf{24.0}              & 970     
&     \textbf{8.9}  & \textbf{10.1}                & 1336               & \textbf{13.8}               & 1913              & \textbf{22.4}              & 1177             \\

\bottomrule
\end{tabular}
}
\caption{Evaluation results on three instruction following benchmarks: AlpacaEval 2, Arena-Hard, and MT-Bench. \textbf{LC} denotes length controlled win rate (\%). \textbf{LWR} denotes length weighted win rate (\%). \textbf{Len.} is the abbreviation of avarege generation length. The best results are highlighted with \textbf{bold}.}
\label{table1}
\end{table*}



\section{Experiments}

\subsection{Experimental Setup}

\textbf{Tasks.}
We evaluated our method on the open-ended instruction-following task, mathematical reasoning task, and the commonsense reasoning question-answering (QA) tasks. 
The experimental evaluation is organized into three main scenarios. First, we evaluate RoPO's alignment capability on instruction-following tasks, following the approach of most existing works. Second, we assess the level of knowledge forgetting in two different settings. Following \citet{dpop}, we first apply supervised fine-tuning (SFT) and DPO on mathematical reasoning tasks. We then evaluate the model on both in-distribution (ID) mathematical reasoning benchmarks and out-of-distribution (OOD) commonsense QA benchmarks, in order to assess its retention of knowledge acquired during the pre-training stage. In addition, we first perform SFT on the commonsense QA training set, followed by SFT and DPO on instruction-following tasks. This setup is designed to evaluate how well the model retains knowledge learned during post-training.




\noindent \textbf{Training details.} We chose the popular Meta-Llama-3-8B \cite{dubey2024llama3herdmodels}, Mistral-7B-v0.1 \cite{jiang2023mistral7b}, and Llama2-13B \cite{touvron2023llama2} as the backbone model. For the open-ended instruction-following task, we supervised fine-tuning (SFT) a base model on the UltraChat-200k dataset \cite{ding-etal-2023-enhancing} to acquire a SFT model. Subsequently, we conduct preference optimization on the UltraFeedback dataset  \cite{cui2024ultrafeedbackboostinglanguagemodels} using the SFT model as the reference model. For the mathematical reasoning task, we supervised fine-tuning (SFT) a base model on the MetaMathFewshot dataset and then conduct preference optimization on MetaMath-DPO-FewShot dataset \cite{dpop}.
For the commonsense reasoning QA tasks, it consist of 8 sub-tasks, each of which is equipped with a predefined training and testing set. We merged these 8 training sets into an integrated training set. For more training details, please refer to Appendix \ref{train}.

\begin{figure}[!t]
  \centerline{\includegraphics[scale=0.24]{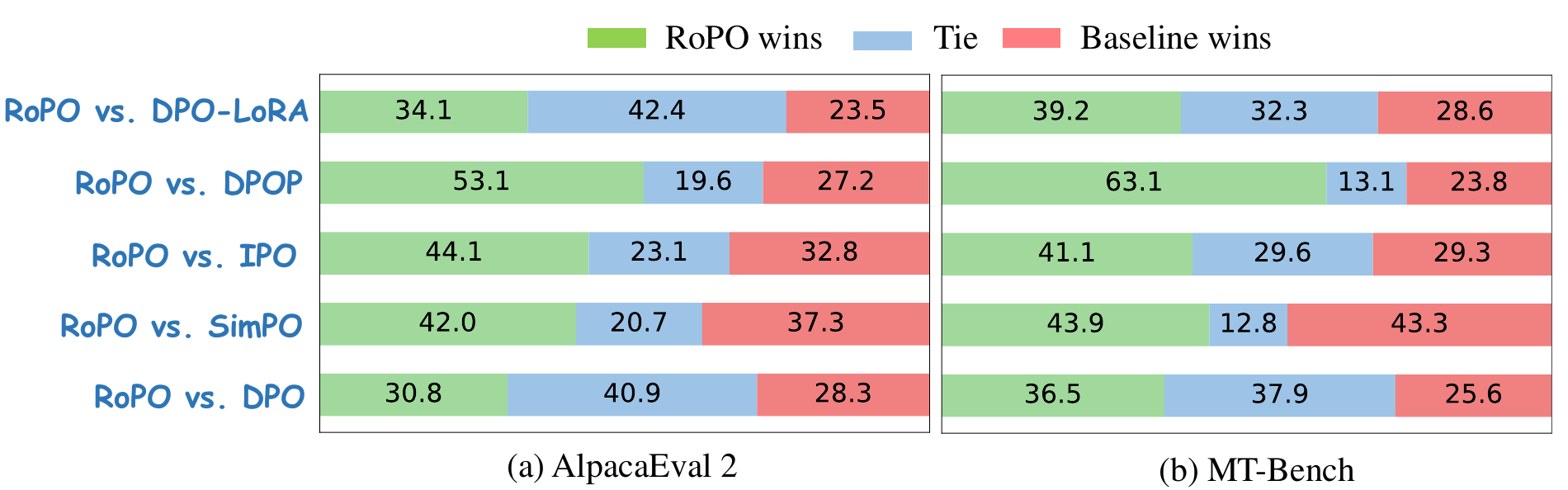}}
  \caption{Human evaluation results on the AlpacaEval 2 and  MT-Bench dataset (Meta-Llama-3-8B). The result is statistically significant with p-value < 0.05, and \textbf{Kappa} ($\kappa$) falls between 0.5 and 0.7. }
  \label{human-res}
\end{figure} 

\noindent \textbf{Baselines.}
We compare RoPO with the following offline preference optimization methods: DPO \cite{rafailov2023directpreferenceoptimizationlanguage}, SimPO \cite{meng2024simposimplepreferenceoptimization}, R-DPO \cite{RDPOpark2024disentanglinglengthqualitydirect}, KTO \cite{KTOethayarajh2024ktomodelalignmentprospect}, WPO \cite{zhou2024wpoenhancingrlhfweighted} DPO* ($\beta=0.3$), IPO \cite{azar2023generaltheoreticalparadigmunderstand}, ORPO \cite{ORPOhong2024orpomonolithicpreferenceoptimization}, DPOP \cite{dpop}, Inter-DPO \cite{kojima2024intermediatedirectpreferenceoptimization}, SamPO \cite{lu2024eliminatingbiasedlengthreliancesampo}, LD-DPO \cite{liu2024lengthdesensitizationdirectpreferencelddpo}, and DPO-LoRA. The DPO* baseline is configured by augmenting the $\beta$ parameter, which regulates the strength of the KL divergence in DPO. More detailed introductions of these baselines are given in the Appendix \ref{baselines}.



\begin{table*}[]
\centering
\setlength{\tabcolsep}{6pt}
\renewcommand\arraystretch{1.0}
\scalebox{0.82}{
\begin{tabular}{lccccccccccc}
\toprule
\multirow{2}{*}{\textbf{Method}} & \multicolumn{1}{|c}{\textbf{\textsc{New Task}\ }} & \multicolumn{9}{|c|}{\textbf{\textsc{Old Task} } }  \\ & \multicolumn{1}{c}{\textbf{AlpacaEval 2}}  & \textbf{BoolQ} & \textbf{PIQA} & \textbf{SIQA} & \textbf{Hella.} & \textbf{Wino.} & \textbf{ARC-e} & \textbf{ARC-c} & \textbf{OBQA} & \textbf{Avg.} \\ \hline
\rowcolor[rgb]{0.957,0.957,0.957}
TS-FT     &  3.4     & 74.2          & 86.7         & 80.5         & 94.0              & 86.0               & 90.2          & {80.5}          & \textbf{88.4}         & {85.0}              \\ 
DPO     &   11.6     &   73.0         & 86.9           &   79.9         &  93.9              & 85.7               &  88.4        &  78.7         &  84.3         & 83.8            \\
SimPO     &    12.7
&  33.6       &  79.4   
&  74.8      & 63.9       
&  80.0       &   81.6  
&  70.3       & 80.0    
&  70.5          \\

R-DPO     &  11.4    &  25.3       &  84.7      &  76.7      & 59.2             &  80.4            &   83.6     &  72.6       & 80.0       &  70.4         \\
KTO     &  8.0   &  44.5     &  85.0     &   72.5     &  93.4         &  78.6      & 82.7     &  72.3            &  80.8       &  76.2       \\
WPO     &   13.5   &  74.0     &  85.8     &   81.0     &  93.9         &  85.4      & 89.6     &  79.9            &  84.6       &  84.3       \\
IPO     &  7.6     &   \textbf{74.9}     &   \textbf{88.0}     & 81.4        &  94.3    &  {86.0}             &  \textbf{90.4}       &  80.7     & 85.0      &  85.1          \\
DPOP     &  9.4   &  63.2     &   86.1       &  80.0     &  93.7        &   74.6       &  89.8         &         78.1    &  85.6        &   81.4       \\
SamPO     &  13.0   &   73.0         & 86.5           &   79.3         &  93.0              & 86.1               &  88.3        &  78.4        &  84.9         & 84.2            \\
LD-DPO     &   12.9   &  68.9     &   80.2       &  78.5     &  89.0        &   82.8       &  85.3         &        76.6    &  83.6        &   80.6       \\
DPO-LoRA     &   9.4   &  74.4     &   \textbf{88.0}       &  80.9     &  \textbf{94.4}        &   86.0       &  90.2         &         80.3    &  85.6        &   85.0       \\
\rowcolor[rgb]{0.87,0.94,1}
RoPO    &    \textbf{13.7}   &  {74.7}           &  \textbf{88.0}         & \textbf{81.5}           &   {94.2}                 &     \textbf{86.3}            &   89.9          &   \textbf{82.0}             &    86.9       &   \textbf{85.4}         \\ \bottomrule
\end{tabular}
}
\caption{Accuracy comparison of aligned \textbf{Meta-Llama-3-8B} model with various methods on 8 commonsense reasoning datasets. TS-FT denotes model fine-tuning on the training set of task-specific datasets.}
\label{table2}
\end{table*}

\begin{table}[!t]
\centering
\renewcommand\arraystretch{1.0}
\setlength{\tabcolsep}{3pt}
\scalebox{0.85}{
\begin{tabular}{lccccccc}
\toprule
\multirow{2}{*}{\textbf{Method}} & \multicolumn{1}{|c}{\textbf{\textsc{MATH}\ (ID)}} & \multicolumn{2}{|c|}{\textbf{\textsc{Commonsense}\ (OOD)}}  \\  & \textbf{GSM8K(5)}  & \textbf{ARC(25)} & \textbf{HellaSwag(10)}                                                             \\ \hline
\rowcolor[rgb]{0.957,0.957,0.957}
SFT             & 75.1      & {58.9}       & 81.3            \\ 
DPO             & \underline{77.3}       & 57.2       & 81.0               \\
SimPO           & \textbf{77.5}      & 56.2   & 80.4          \\
WPO           & {77.2}      & 56.1   & 80.7          \\
R-DPO           & {75.4}   & {54.1}   & 79.3          \\
IPO             & {76.0}   & \textbf{58.7}   & \underline{81.1}       \\
DPOP             & {76.9}   & {57.6}   & 80.9       \\
Inter-DPO            & {74.3}   & {58.0}   & 80.3         \\
SamPO            & {77.0}   & {58.4}   & 80.6         \\
LD-DPO            & {76.5}   & {57.8}   & 80.0         \\
DPO-LoRA            & {76.8}   & \underline{58.2}   & \underline{81.1}         \\
\rowcolor[rgb]{0.87,0.94,1}
RoPO         & \underline{77.3}            & \textbf{58.7}   & \textbf{81.3}      
\\
 \bottomrule
\end{tabular}
}
\caption{Evaluation results on GSM8K, ARC, and HellaSwag tasks on the Meta-Llama-3-8B model.}
\label{tablemath}
\end{table}

\noindent \textbf{Evaluation.}
 We employed three challenging benchmarks to evaluate the open-ended instruction-following task: MT-Bench \cite{zheng2023judgingllmasajudgemtbenchchatbot}, AlpacaEval 2 \cite{alpaca_eval}, and Arena-Hard \cite{arenahardli2024crowdsourceddatahighqualitybenchmarks}. 
In addition, we also evaluate on the commonsense reasoning QA task after training on the instruction-following datasets to assess knowledge forgetting. 
We evaluate on the GSM8K (ID) \cite{cobbe2021trainingverifierssolvemath}, ARC (OOD) \cite{clark2018thinksolvedquestionanswering}, and HellaSwag (OOD) \cite{zellers-etal-2019-hellaswag} datasets when training on MetaMath. For the automatic evaluation metric, we utilize the length-controlled (\textbf{LC}) win rate and length-weighted win rate (\textbf{LWR}) against the GPT-4 to avoid the length bias issue in GPT-4 evaluation ( Equation \ref{eqlwr}). In addition, to further mitigate the length bias inherent in GPT-based evaluations, we conducted human evaluations. Five well-educated annotators were asked to choose the superior response based on evaluation instructions in Figure \ref{human-eval-ins}.
Further details of evaluation are provided in Appendix \ref{eval}. 




\subsection{Experimental Results}

\subsubsection{Main Results}
\textbf{RoPO consistently outperforms all the preference optimization baselines across instruction-following benchmarks.} Table \ref{table1} shows that RoPO achieves state-of-the-art performance on LC and LWR while maintaining shorter generation lengths, effectively alleviating the verbosity issue commonly caused by DPO. 
As shown in Figure \ref{fig4}, RoPO demonstrates consistently strong performance on the one-turn questions of MT-Bench, excelling in role-play, reasoning, extraction, coding, and writing tasks. The human evaluations results in Figure \ref{human-res} also show that RoPO fulfills instruction requirements with shorter generations.



\begin{table*}[]
\centering
\renewcommand\arraystretch{1.0}
\setlength{\tabcolsep}{3pt}
\scalebox{0.84}{
\begin{tabular}{lccccccccccccccc}
\toprule
\multirow{3}{*}{\textbf{Method}} & \multicolumn{7}{c}{\textbf{Mistral-Base (7B)}}                                                                              & \multicolumn{7}{c}{\textbf{Llama3-Base (8B)}}                                                                               \\ \cline{2-15} 
                                 & \multicolumn{3}{c}{\textbf{AlpacaEval 2}} & \multicolumn{2}{c}{\textbf{Arena-Hard}} & \multicolumn{2}{c}{\textbf{MT-Bench}} & \multicolumn{3}{c}{\textbf{AlpacaEval 2}} & \multicolumn{2}{c}{\textbf{Arena-Hard}} & \multicolumn{2}{c}{\textbf{MT-Bench}} \\ \cline{2-15} 
                              & \textbf{LC}     & \textbf{LWR}          & \textbf{Len.}      & \textbf{LWR}         & \textbf{Len.}     & \textbf{LWR}        & \textbf{Len.}   & \textbf{LC}   & \textbf{LWR}          & \textbf{Len.}      & \textbf{LWR}         & \textbf{Len.}     & \textbf{LWR}        & \textbf{Len.}    \\ \hline
                                 \rowcolor[rgb]{0.87,0.94,1}
RoPO                 &   \textbf{11.4}    & \textbf{13.5}                & 1042               & \textbf{16.0}               & 1433              & \textbf{23.6}              & 970        &   \textbf{13.7}      & \textbf{11.9}                & 1083               & \textbf{18.3}               & 1516              & \textbf{28.3}              & 1027             \\
w/o GRM      &  9.5                  
& 10.8                 & 939 
& 13.2               & 1274     
& 19.6              & 946  
&   10.6
& 9.4                 & 917    
& 14.4               & 1465  
& 16.8              & 931              \\
w/o FRM      &    10.2             
& 11.7                & 1070 
& 13.5               & 1459  
& {21.8}     & 1014     
& 11.4
& 9.9                 & 1092  
& 16.9               & 1526  
& 22.6              & 993              \\
DPO-DoRA$ _{r=4}$             &   10.0    & 10.6                & 967                & 14.9               & 1428              & 17.0              & 953       &   11.2    & 9.0                 & 998                & 15.5               & 1513              & 20.7              & 994              \\

\bottomrule
\end{tabular}
}
\caption{Ablation experiment results on three instruction-following benchmarks.}
\label{table3}
\end{table*}

\textbf{RoPO effectively mitigates reward hacking issue in preference optimization algorithms.} As previously observed, RoPO demonstrates superior performance with reasonably constrained generation length, suggesting that it effectively addresses the issue of verbosity. We now analyze RoPO’s ability to alleviate reward hacking from the perspective of knowledge forgetting and repetition. Table \ref{table2} shows that, in the continual learning setting, RoPO achieves the best performance on the new instruction-following task while retaining strong results on the previous QA tasks, with improved accuracy across six datasets. We speculate that this is because RoPO better preserves the knowledge encoded in the angular relationships between neurons, while also acquiring additional commonsense knowledge during the alignment process. Table \ref{tablemath} shows that when mathematics is the target training task, RoPO achieves the second-best performance among all preference optimization methods on GSM8K, while outperforming all baselines on the out-of-distribution (OOD) commonsense knowledge datasets. Regarding generation repetitiveness, RoPO achieves Pareto optimal performance across the LWR and diversity metrics. In contrast, all baselines except IPO reduce the diversity of generated content significantly, as illustrated in Figure \ref{scatter}. The overall findings indicate that RoPO effectively mitigates the reward hacking issue while maintaining strong preference learning performance.

\subsubsection{Ablation Study}
To evaluate the efficacy of RoPO, we carried out ablation studies. In Table \ref{table3}, we present results from ablating each key design of RoPO: (1) removing the global rotation matrix $\bf{H}$ (i.e. {w/o GRM}); (2) removing the fine-trained rotation matrix $\tilde{\bm{G_1}}\tilde{\bm{G_2}}$ (i.e. {w/o FRM}); (3) removing the rotation design in the directional matrix, and degradating to DoRA \cite{liu2024doraweightdecomposedlowrankadaptation}. The results illustrate that every design of RoPO is crucial as eliminating each design would result in varying degrees of performance degradation, indicating that the design of multi-granularity orthogonal matrix enhances the outcomes of optimization process.
  DoRA could be regarded as one version of weight regularization without retaining relative angular distances between weight vectors constant. Experimental results shows that although DoRA also curbed the over-expansion of length bias, its instruct-following ability declines significantly. 

\begin{figure}[!t]
  \centerline{\includegraphics[scale=0.33]{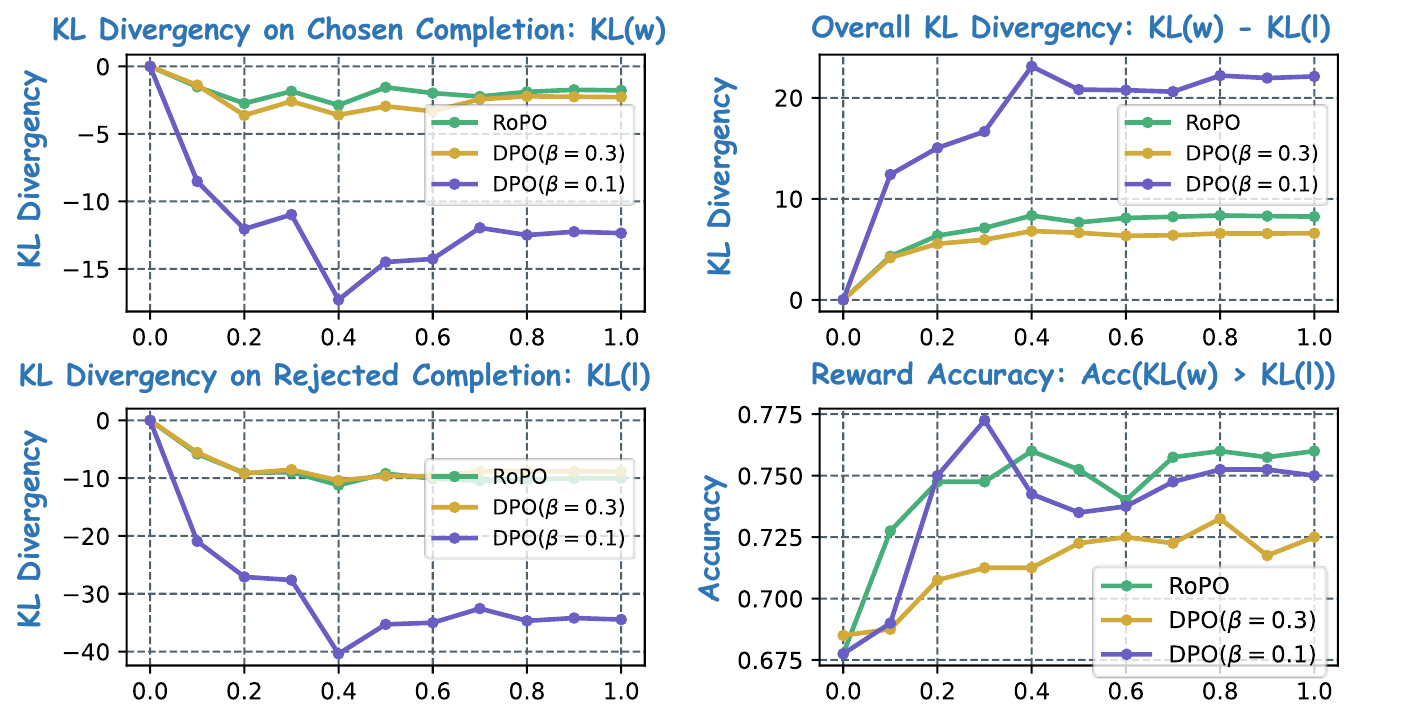}}
  \caption{Comparison of KL divergences and reward accuracy evolution
over 1 epoch during training on the UltraChat-200k (Meta-Llama-3-8B). }
  \label{final-kl}
\end{figure}

\subsection{Qualitative Examples}

To intuitively demonstrate the effectiveness of RoPO, we compare the responses generated by RoPO with different baselines. The examples presented in Table \ref{table5} and Table \ref{table6} show that reward overoptimization could significantly degrades the model's performance on simple tasks, resulting in repetitive or irrelevant generations human dislike. RoPO effectively mitigates the degeneration phenomenon. Table \ref{table7} showcases a more challenging mathematical problem. SimPO's response was highly disorganized, while DPO exhibited a commonsense hallucination by suggesting that the probability of an event could exceed $1$. In contrast, although RoPO's answer was incorrect, its reasoning process was consistent with that of GPT-4. Table \ref{table8} and Table \ref{table9} highlights that RoPO possesses outstanding open-domain generation capability.



\subsection{Analysis}

In this section, we will delve deeper into RoPO's underlying mechanisms. 
The variations depicted in Figure \ref{final-kl} indicate that RoPO exhibits a similar trend to DPO*, effectively preventing the excessive suppression of generation probabilities on both chosen and rejected completions, while maintaining reward accuracy on par with DPO. Additionally, its convergence curve is smoother and more stable, further validating the effectiveness of the regularization. We also conducted a visual analysis of the neuron arrangement using the t-SNE technique \cite{JMLR:v9:vandermaaten08a}. As shown in Figure \ref{figure10}, DPO training leads to a disruption in the isotropy of sampled neurons, whereas RoPO maintains the integrity of the neuron arrangement.



\section{Conclusion}

In this paper, We delve into the underlying causes behind DPO reward hacking from a fresh parameter perspective. Based on the analysis, we proposed RoPO, which implicitly constrains the output layer logits with the KL divergence inherited from DPO and explicitly constrains the intermediate hidden states by fine-tuning on a multi-granularity orthogonal matrix. Experiments on extensive benchmarks demonstrate that RoPO not only enhances alignment performance but also avoids reward hacking. 


\section*{Limitations}
RoPO effectively addresses reward hacking problem in DPO. However, we acknowledge certain limitations in our work. 

(1) Currently, we have only applied the RoPO constraint to the query and value vectors within the attention layers. In the future, we intend to extend this approach to additional layers.

(2) In the analysis section, given that some baselines do not have a reference model, we did not plot their KL divergence changes. We will do further analyses on these baselines in the future. 

(3) Due to limitations in resources and time, we were unable to validate the generalizability of our method on more and larger models. In the future, we plan to evaluate our approach on a wider range of models.

\section*{Ethics Statement}

The benchmark datasets we utilized in our experiments are all highly respected, open-source datasets. The backbone LLMs are also open-sourced. They were compiled with strict adherence to user privacy protection protocols, ensuring the exclusion of any personal information. Moreover, our proposed approach is conscientiously designed to uphold ethical standards and promote societal fairness, guaranteeing that no bias is introduced. 
We use AI writing in our work, but only for polishing articles to enhance their readability.

\section*{Acknowledgments}

This work was supported by the National Natural Science
Foundation of China (No. 62472419, 62472420).

\bibliography{anthology,custom}
\bibliographystyle{acl_natbib}

\appendix

\section{More Implementation Details}
\label{train}
We discover that hyperparameter tuning is of paramount significance for attaining the optimal performance of preference optimization approaches. Hence, to acquire the supreme performance, we executed a sophisticated hyperparameter search. Below, we exhibit the hyperparameter configurations in the experiment. 

Regarding the SFT training, we train models by utilizing the UltraChat-200k dataset with the subsequent hyperparameters: a learning rate of 1e-6, a batch size of 128, a maximum sequence length of 2048, and a cosine learning rate schedule with 10\% warmup steps for 1 epoch. All the models are trained with an Adam optimizer.

For the preference optimization stage, we train the SFT models using the UltraFeedback dataset with the same hyperparameters as SFT training under the full-parameter settings .  The learning rate was set as 2e-5 for DPO-LoRA and DPO-DoRA, 1e-3 for RoPO.  

For the commonsense reasoning QA task, we conducted experiment on it aimed at evaluating reward hacking. we first trained a base model on this commonsense reasoning QA training set until convergence, and then conducted SFT and preference optimization on the instruction-following dataset. The reward hacking problem could be manifested by comparing the performance on the testing set, as it leads to knowledge forgetting.

For the training framework, we implement our experiments on the open-sourced unified LlamaFactory \cite{zheng2024llamafactory}.

\section{More details of Baselines} \label{baselines}
DPO: \citet{rafailov2023directpreferenceoptimizationlanguage} derived it by fitting an implicit reward function through the reparameterization.

IPO: \citet{azar2023generaltheoreticalparadigmunderstand} revised the objective to minimize the disparity between the ratio of log-likelihoods and a given threshold to mitigate the overfitting problem of DPO.

KTO: \citet{KTOethayarajh2024ktomodelalignmentprospect} proposed it to directly maximize the utility of generations instead of maximizing the log-likelihood of preferences.

ORPO: \citet{ORPOhong2024orpomonolithicpreferenceoptimization} integrates a penalty term to preclude the learning of undesirable responses while augmenting the probability of learning preferred ones.

R-DPO: \citet{RDPOpark2024disentanglinglengthqualitydirect} attempted to add a length regularization term in the loss function to alleviate the abnormally long generation issue.

SimPO: \citet{meng2024simposimplepreferenceoptimization} removed the reference model and used the average log probability of a sequence as the implicit reward.

WPO:  \citet{zhou2024wpoenhancingrlhfweighted} focuses on improving alignment performance by simulating on-policy preference data using off-policy preference data.

Inter-DPO: \citet{kojima2024intermediatedirectpreferenceoptimization} revised the DPO loss as the weighted sum of the final-layer DPO and intermediate DPO losses.

SamPO: \citet{lu2024eliminatingbiasedlengthreliancesampo} identifies an algorithmic bias toward length reliance in DPO, where longer responses receive disproportionately larger gradient updates. Their solution involves down-sampling to eliminate this length dependence.

LD-DPO: \citet{liu2024lengthdesensitizationdirectpreferencelddpo} attributed DPO's length sensitivity to how text length affects likelihood and proposed to decompose the response likelihood into public-length and excessive-length components, then reducing weights for the latter to mitigate verbosity preferences.

To acquire the supreme performance of each baseline, We conducted a parameter search based on the search space provided in the SimPO paper and the papers introducing the baselines. The following hyperparameter configuration was selected based on the best results obtained on the validation set: DPO: $\beta=0.1/0.3$, IPO: $\tau=2.0$, KTO: $\lambda_l=\lambda_w=1.0,\beta=0.1$, ORPO: $\lambda=0.1$, R-DPO: $\alpha=0.003,\beta=0.1$, SimPO: $\beta=2.0, \gamma=0.5$, DPOP: $\beta=0.1, \lambda=5$, Inter-DPO: $\beta=0.1, \gamma=0.9$, WPO: $\beta=0.1$, SamPO: $\beta=0.1$, LD-DPO: $\beta=0.1, \alpha=0.3$,

\section{More Evaluation Setups} \label{eval}

We employed three challenging benchmarks to evaluate the instruction-following task: MT-Bench \cite{zheng2023judgingllmasajudgemtbenchchatbot}, AlpacaEval 2 \cite{alpaca_eval}, and Arena-Hard \cite{arenahardli2024crowdsourceddatahighqualitybenchmarks}. Among them,
AlpacaEval 2 consists of 805 questions. MT-Bench contains 80 questions falling into the following eight common categories: writing, role-play, extraction, reasoning, math, coding, knowledge I (STEM), and knowledge II (humanities/social science). Arena-Hard is an extension of MT-Bench, which collects 500 challenging high-quality prompts, achieving a state-of-the-art agreement with human preference rankings. For decoding hyperparameters in evaluation, we use a sampling decoding strategy to generate responses, with a temperature of 0.95, top-p of 0.7, and top-k of 50.

We present the findings using automatic and LLM-based evaluation methods. For LLM-based evaluation, we employ gpt-4o-2024-08-06  
as the judge model to conduct pairwise comparisons for each preference optimization method and GPT-4. We consider the win rate (WR) against the responses generated by GPT-4 on AlpacaEval 2, MT-Bench, and Arena-Hard. We provide the prompt for evaluation in Table. To alleviate positional bias, we assess each candidate in both positions within two separate runs, and the ultimate result is calculated as the average of the two runs. For MT-Bench, we additionally report the average MT-Bench score with gpt-4o-2024-08-06.  Since GPT-4 has a tendency to give higher scores to longer responses during evaluation \cite{dubois2024lengthcontrol}, the length should be taken into account simultaneously when comparing performances. Under similar scores, we hold that shorter responses are superior. Therefore, we report the length-weighted win rate (LWR) as the final results by multiplying with the ratio of the content length generated by GPT-4 to that generated by the current method to exclude the influence of length as shown in shown in the following formula. 
\begin{equation}
\text{LWR}= \text{WR} \cdot \frac{\text{Len}(y_\text{g4})}{\text{Len}(y)}
\label{eqlwr}
\end{equation}
where $y_\text{g4}$ denotes responses generated by GPT-4, and Len denotes length of generation.

For automatic evaluation, given that the reward over-optimization in alignment process causes damage to the diversity of the generated content, we add the corresponding metrics to assess it. We utilize the Distinct N-grams metric to evaluate the generation diversity\cite{li-etal-2016-diversity} through calculating a geometric mean of Distinct N with $N = 1,2,3,4$. For commonsense reasoning QA, we regard accuracy as the evaluation metric across 8 test sets.

\begin{equation}
\text{Diversity}=\sqrt[4]{\prod\limits_{n=1}^{4} \text{Distinct-n}}.
\end{equation}

\begin{figure}[htbp]
  \centerline{\includegraphics[scale=0.7]{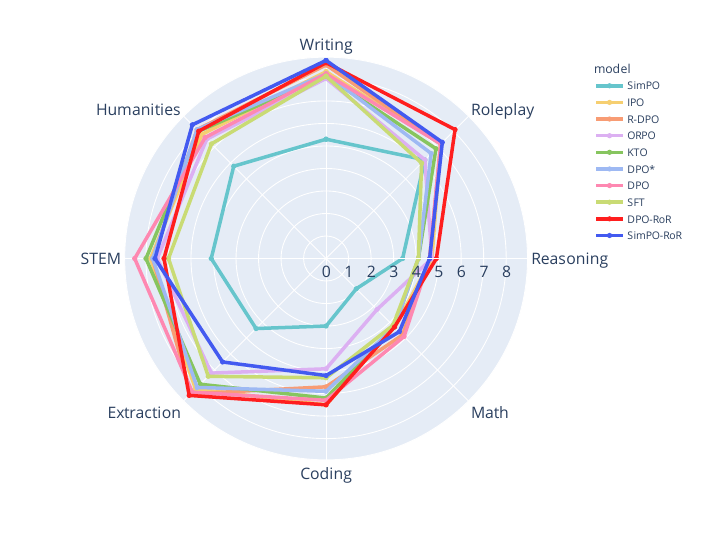}}
  \caption{Scores of RoPO compared with baselines in MT-Bench on the Meta-Llama-3-8B backbone.}
  \label{fig4}
\end{figure} 


\begin{figure}[!t]
  \centerline{\includegraphics[scale=0.46]{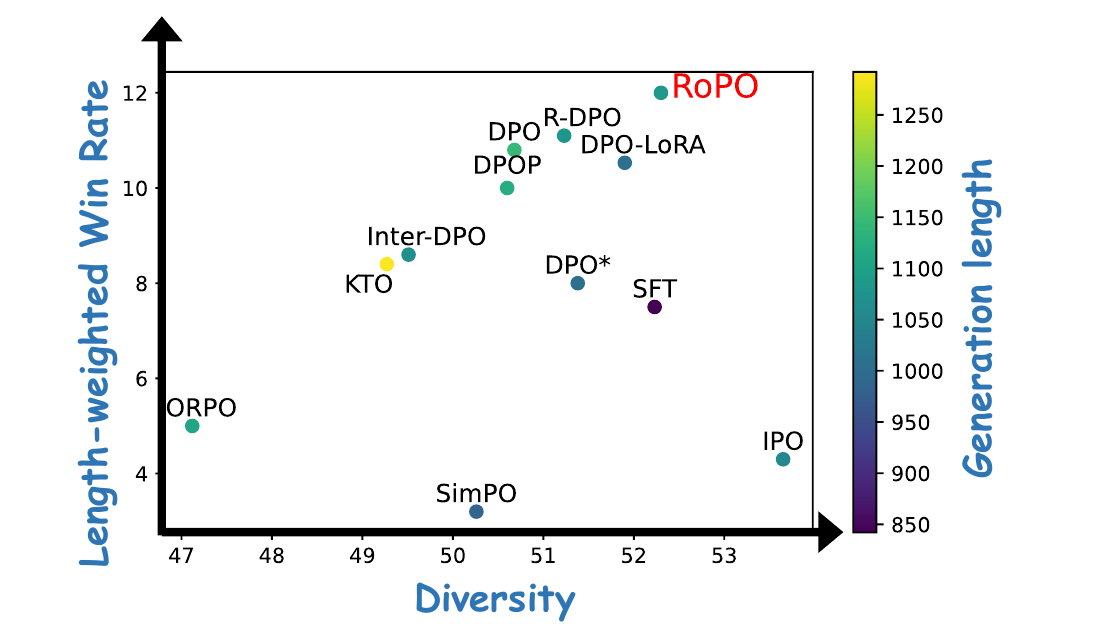}}
  \caption{experimental results on Llama3-8B. }
  \label{scatter}
\end{figure}

\section{More Analyses of Knowledge Retaining}

To validate the issue of knowledge forgetting caused by alignment overfitting, we compare the influence of different preference optimization methods on the performance of commonsense reasoning tasks. Table \ref{table2} exhibits that DPO, KTO, and R-DPO resulted in a decline of the model's general ability. By observing some response cases, we discovered that the model appeared to directly answer the content of the options instead of answering the options themselves. It even provided safe response like "Sorry, I don't know." This implies that overfitting during the alignment training lead to the forgetting of task format knowledge and the decline of the model's question understanding ability. In contrast, RoPO still maintain the performance well on Commonsense Reasoning QA, with the accuracy rate on 6 datasets enhanced. We speculate that this is because RoPO retains the knowledge encoded in the angle between neurons well and acquires additional commonsense knowledge during the alignment process.

\section{$\beta$ in KL divergency.}
In the DPO objective, $\beta$ governs the deviation from the reference model $\pi_{\mathrm{ref}}$. We believe that through adjusting $\beta$, the alignment intensity can be regulated, thereby controlling reward hacking. The following is our explanation. Assuming that the true preference of a sample $p^*(y_w\succ y_l|x)=\hat{p}$, if the $\beta$ value shrinks to  approach $0$, the model has to learn to further reduce the probability of rejected completions $\pi_\theta(y_l| x)$ to fit the true preference probability $\hat{p}$; if the $\beta$ value increases, the model merely needs to learn to decrease the probability of rejected completions with a relatively weaker strength.

\section{Proof of Orthogonality} \label{Ortho}

\subsection*{Definition of Householder Reflection Matrix}

Given a non-zero vector $\mathbf{v} \in \mathbb{R}^n$, the corresponding Householder reflection matrix $H$ is defined as:

$$
H = I - 2 \frac{\mathbf{v} \mathbf{v}^T}{\mathbf{v}^T \mathbf{v}},
$$

where $I$ is the identity matrix and $\mathbf{v} \mathbf{v}^T$ is the outer product matrix.

\subsection*{Properties of Householder Matrices}

\begin{enumerate}
    \item \textbf{Symmetry:} $H^T = H$. \\
    \textit{Proof:}
    \begin{equation}
    \begin{aligned}
    H^T =& \left( I - 2 \frac{\mathbf{v} \mathbf{v}^T}{\mathbf{v}^T \mathbf{v}} \right)^T 
    \\=& I^T - 2 \frac{(\mathbf{v} \mathbf{v}^T)^T}{\mathbf{v}^T \mathbf{v}} \\
    =& I - 2 \frac{\mathbf{v} \mathbf{v}^T}{\mathbf{v}^T \mathbf{v}} = H.
    \end{aligned}
    \end{equation}

    \item \textbf{Orthogonality:} $H^T H = I$. \\
    \textit{Proof:}
    $$
    H^T H = H H \quad (\text{since } H \text{ is symmetric})
    $$
    \begin{equation}
    \begin{aligned}
    H H =& \left( I - 2 \frac{\mathbf{v} \mathbf{v}^T}{\mathbf{v}^T \mathbf{v}} \right) \left( I - 2 \frac{\mathbf{v} \mathbf{v}^T}{\mathbf{v}^T \mathbf{v}} \right) \\
    =& I - 4 \frac{\mathbf{v} \mathbf{v}^T}{\mathbf{v}^T \mathbf{v}} + 4 \frac{\mathbf{v} \mathbf{v}^T \mathbf{v} \mathbf{v}^T}{(\mathbf{v}^T \mathbf{v})^2}.
    \end{aligned}
    \end{equation}
    Since $\mathbf{v}^T \mathbf{v}$ is a scalar, we have:
    $$
    \mathbf{v} \mathbf{v}^T \mathbf{v} \mathbf{v}^T = \mathbf{v} (\mathbf{v}^T \mathbf{v}) \mathbf{v}^T = (\mathbf{v}^T \mathbf{v}) \mathbf{v} \mathbf{v}^T.
    $$
    Therefore:
    \begin{equation}
    \begin{aligned}
    H H =& I - 4 \frac{\mathbf{v} \mathbf{v}^T}{\mathbf{v}^T \mathbf{v}} + 4 \frac{(\mathbf{v}^T \mathbf{v}) \mathbf{v} \mathbf{v}^T}{(\mathbf{v}^T \mathbf{v})^2} \\ =& I - 4 \frac{\mathbf{v} \mathbf{v}^T}{\mathbf{v}^T \mathbf{v}} + 4 \frac{\mathbf{v} \mathbf{v}^T}{\mathbf{v}^T \mathbf{v}} = I.
    \end{aligned}
    \end{equation}
    Thus, $H^T H = I$, proving that $H$ is orthogonal.
\end{enumerate}

\subsection*{Product of Two Householder Matrices}

Let $H_1$ and $H_2$ be two Householder matrices:
$$
H_1 = I - 2 \frac{\mathbf{v}_1 \mathbf{v}_1^T}{\mathbf{v}_1^T \mathbf{v}_1}, \quad H_2 = I - 2 \frac{\mathbf{v}_2 \mathbf{v}_2^T}{\mathbf{v}_2^T \mathbf{v}_2}.
$$
We want to show that $\tilde{H} = H_1 H_2$ is orthogonal, i.e., $\tilde{H}^T \tilde{H} = I$.

\begin{enumerate}
    \item Compute $\tilde{H}^T$:
    $$
    \tilde{H}^T = (H_1 H_2)^T = H_2^T H_1^T.
    $$
    Since $H_1$ and $H_2$ are symmetric ($H_1^T = H_1$, $H_2^T = H_2$):
    $$
    \tilde{H}^T = H_2 H_1.
    $$
    
    \item Compute $\tilde{H}^T \tilde{H}$:
    $$
    \tilde{H}^T \tilde{H} = (H_2 H_1)(H_1 H_2) = H_2 (H_1 H_1) H_2.
    $$
    Because $H_1$ is orthogonal ($H_1 H_1 = I$):
    $$
    \tilde{H}^T \tilde{H} = H_2 I H_2 = H_2 H_2.
    $$
    Since $H_2$ is orthogonal ($H_2 H_2 = I$):
    $$
    \tilde{H}^T \tilde{H} = I.
    $$
    Therefore, $\tilde{H} = H_1 H_2$ is orthogonal.
\end{enumerate}

\subsection*{Definition of Rotation Matrix}

A rotation matrix $G \in \mathbb{R}^{n \times n}$ is a matrix that performs a rotation in Euclidean space. In $\mathbb{R}^2$, the rotation matrix by angle $\theta$ is:

$$
G(\theta) = \begin{bmatrix}
\cos\theta & -\sin\theta \\
\sin\theta & \cos\theta
\end{bmatrix}
$$

\subsection*{Properties of Givens Matrices}

\begin{enumerate}
    \item Compute the transpose:
    $$
    G(\theta)^T = \begin{bmatrix}
    \cos\theta & \sin\theta \\
    -\sin\theta & \cos\theta
    \end{bmatrix}
    $$
    
    \item Verify $R^T R = I$:
$$
\makebox[\linewidth][c]{%
\scalebox{0.7}{%
\begin{minipage}{0.6\textwidth} 
\centering
\begin{equation*}
    \begin{aligned}
    &G^T G = \begin{bmatrix}
    \cos\theta & \sin\theta \\
    -\sin\theta & \cos\theta
    \end{bmatrix}
    \begin{bmatrix}
    \cos\theta & -\sin\theta \\
    \sin\theta & \cos\theta
    \end{bmatrix}
    \\&= \begin{bmatrix}
    \cos^2\theta + \sin^2\theta & -\cos\theta\sin\theta + \sin\theta\cos\theta \\
    -\sin\theta\cos\theta + \cos\theta\sin\theta & \sin^2\theta + \cos^2\theta
    \end{bmatrix}
    \end{aligned}
\end{equation*}
\end{minipage}%
}%
}
$$

    Using trigonometric identities:
    $$
    = \begin{bmatrix}
    1 & 0 \\
    0 & 1
    \end{bmatrix} = I
    $$
    
    \item Similarly, $G G^T = I$ can be verified.
\end{enumerate}

Since the product of orthogonal matrices still satisfies orthogonality, the Multi-Granularity Orthogonal Matrix possesses orthogonality.

\section{Hyperspherical Energy} \label{hyper}
The initial proposal of Hyperspherical Energy (HE) was motivated by the diversification and balanced arrangement of neurons to prevent the parameter redundancy problem \cite{liu2020learningminimumhypersphericalenergy}. Inspired by the renowned physics problem known as Thomson problem, \citet{liu2020learningminimumhypersphericalenergy} designed the neural network training objective with Minimum Hyperspherical Energy (MHE) as the regularization. Assuming that there is a fully connected layer $\boldsymbol{W=}\{\boldsymbol{w}_1,\cdots,\boldsymbol{w}_n\}\boldsymbol{\in}\mathbb{R}^{d\times n}$, where $\boldsymbol{w}_i\in\mathbb{R}^d$ denotes the $i$-th neuron. The definition of HE is as follows:
\begin{equation}
{\mathrm{HE}}(\boldsymbol{W}) = \sum_{i\neq j}\|\hat{\boldsymbol{w}}_i-\hat{\boldsymbol{w}}_j\|^{-1}
\label{he1}
\end{equation}
where $\boldsymbol{\hat{w}}_i=\boldsymbol{w}_i/\|\boldsymbol{w}_i\|$ is the $i$-th normalized neuron.

\begin{equation}
{\Delta\mathrm{HE}}(\boldsymbol{W}) = {\mathrm{HE}}(\boldsymbol{W'})-{\mathrm{HE}}(\boldsymbol{W})
\label{dhe}
\end{equation}
where $\boldsymbol{W'}$ denotes the aligned model, and $\boldsymbol{W}$ denotes the model before alinging.

Our RoPO satisfies the following equation:
\begin{equation}
\sum_{i\neq j}\|\hat{\boldsymbol{w}}_i-\hat{\boldsymbol{w}}_j\|^{-1}-\sum_{i\neq j}\|\hat{\boldsymbol{w}}_i^0-\hat{\boldsymbol{w}}_j^0\|^{-1}=0,
\end{equation}
where $\hat{\boldsymbol{w}}_i^0$ denotes the weight before fine-tuning, and $\hat{\boldsymbol{w}}_i$ denotes the weight after fine-tuning.


\section{Training Efficiency}

In addition to its outstanding comprehensive performance, RoPO also has the advantages of low trainable parameters. Compared with other preference optimization baselines ($100\%$ parameters), RoPO merely demands 0.0151\% of the trainable parameters. Next, we conduct a mathematical analysis: 

Supposing the weight matrix of the neural network is $\boldsymbol{W}\boldsymbol{\in}\mathbb{R}^{d\times n}$, there are three components of trainable parameters in RoPO: magnitude vector $\bf{m}$ with $n$ trainable parameters, fine-grained rotation matrices with $d-1$ trainable parameters, and global rotation matrix with $2d$ trainable parameters. Therefore, the training parameter quantity of RoPO is $3d - 1 + n$. By contrast, the training parameter quantity of the baseline DPO-LoRA is $r \times (d + n)$. In our experimental setup, we apply the trainable matrix to the query vectors and value vectors in the attention mechanism. Assuming that the backbone model employs the common Multi-Head Attention ($d = n$), then the trainable parameter quantity of RoPO is approximately $4d-1$, the parameter quantity of DPO-LoRA (r = 4) and DPO-DoRA (r = 4) is $8d$. RoPO achieves performance exceeding that of DPO-LoRA and DPO-DoRA with fewer parameters.

\begin{figure*}[htbp]
    \centering
\begin{equation*}
\begin{gathered}
\mathbf{R}\cdot\mathbf{x}_m=\begin{bmatrix}x_0\\x_1\\x_2\\x_3\\\vdots\\x_{d-2}\\x_{d-1}\end{bmatrix}\otimes\begin{bmatrix}\cos m\theta_0\\\cos m\theta_0\\\cos m\theta_1\\\cos m\theta_1\\\cos m\theta_1\\\vdots\\\cos m\theta_{d/2-1}\\\cos m\theta_{d/2-1}\end{bmatrix}+\begin{bmatrix}x_1\\x_0\\x_3\\x_2\\\vdots\\x_{d-1}\\x_{d-2}\end{bmatrix}\otimes\begin{bmatrix}-\sin m\theta_0\\\sin m\theta_0\\-\sin m\theta_1\\\sin m\theta_1\\\vdots\\-\sin m\theta_{d/2-1}\\\sin m\theta_{d/2-1}\end{bmatrix}
\end{gathered}
\end{equation*}
\caption{The sparse matrix multiplication acceleration method for Fine-grained Rotation Matrices.}
\label{accelerate}
\end{figure*}

\section{Sparse Matrix Multiplication Implementation}
\label{accele}
Due to the sparsity of Fine-grained Rotation Matrix $\tilde{G_1}$, the matrix multiplication between it and the parameter matrix can be quickly implemented in the equivalent way displayed in Figure \ref{accelerate}.

\section{Proof of Full-Angle Coverage for the Fine-Grained Rotation Matrix} 
\label{frm}

Given any vector $\boldsymbol{v}\in\mathbb{R}^d $, $\tilde{\bm{G_1}}\tilde{\bm{G_2}}$ could rotate it to any vector $\boldsymbol{y}\in\mathbb{R}^d$ on the same sphere with $\boldsymbol{v}$, \textit{i.e.} $\tilde{\bm{G_1}}\tilde{\bm{G_2}}\bm{v}=\bm{y}$.
The theorem ensures that Fine-grained Rotation Matrix (FRM) in RoPO could accomplish any d-dimensional rotation for n neurons synchronously in the weight matrix. Below is our proof:

We assume that the original neuron $\boldsymbol{v}=[v_0,v_1,\dots,v_{d-1}]$ and the target neuron $\boldsymbol{y}=[y_0,y_1,\dots,y_{d-1}]$. Since the magnitude of the vector remains unchanged before and after rotation, we have $||\boldsymbol{v}||^2_2=||\boldsymbol{y}||^2_2$. Without loss of generality, we could normalize the vectors $\boldsymbol{v}$ and $\boldsymbol{y}$ to obtain unit vectors $\hat{\boldsymbol{v}}$ and $\hat{\boldsymbol{y}}$, so that their directions are preserved and the problem reduces to rotating one unit vector to another. In the construction of the FRM, for any two dimensions, there exists a corresponding Givens rotation matrix that performs a rotation in this plane. That is, there are $d-1$ independent parameters to describe this transformation, ensuring a $d-1$ degree of freedom for rotation. Then, we can sequentially rotate the vector to set its dimensions from the $0$-th to $d-2$-th to $0$. Since $\hat{\boldsymbol{v}}$ is a unit vector, its last dimension must be $1$. Next we apply the inverse of the above process to rotate the vector $[0,0,\dots,1]_d$ back to $\hat{\boldsymbol{y}}$.
\begin{equation}
\begin{aligned}
\tilde{\bm{G_1}}&=\prod_{k=0}^{(d/2)-1}\bm{G}(2k,2k+1;\theta_k), \\
\tilde{\bm{G_2}}&=\prod_{k=0}^{(d-1)/2}\bm{G}(2k+1,2k+2;\theta_k),
\end{aligned}
\end{equation}

\section{More Related Work}
\subsection{Parameter-Efficient Fine-Tuning.}
As the scale of the model continues to expand, conducting full fine-tuning of the pre-trained model on downstream tasks is becoming increasingly challenging. The proposal of Parameter-Efficient Fine-Tuning (PEFT) technology has substantially reduced the training and storage costs \cite{gu-etal-2024-light}, significantly expediting the pace of AI research. Currently, there exist three mainstream PEFT approaches \cite{lialin2023scalingscaleupguide}: The first one is adapter tuning, which is accomplished by inserting additional trainable modules into the original model \cite{houlsby2019parameterefficienttransferlearningnlp,liu2022fewshotparameterefficientfinetuningbetter,wang2025sibylempoweringempatheticdialogue}; the second one is prompt tuning, which is achieved by concatenating learnable prefix tokens at the beginning of the input \cite{lester2021powerscaleparameterefficientprompt,liu-etal-2022-p}; the last one is reparameterization tuning, where only the delta of partial model parameters are reparameterized with few trainable parameters \cite{hu2021loralowrankadaptationlarge,aghajanyan2020intrinsicdimensionalityexplainseffectiveness}. The last method can merge the learned delta of model parameters into the original model parameters during inference and does not introduce additional costs at the inference stage. It is commonly held that there are two types of schemes for reparameterization tuning, including LoRA-based methods \cite{gao2024parameterefficientfinetuningdiscretefourier,gu2025adaptoncethriveupdates,liu2024doraweightdecomposedlowrankadaptation,gu2025beamlorabeamconstraintlowrankadaptation} and orthogonal fine-tuning (OFT) \cite{qiu2024controllingtexttoimagediffusionorthogonal,liu2024parameterefficientorthogonalfinetuningbutterfly,GOFTma2024parameterefficientquasiorthogonalfinetuning,chen2024multitaskroleplayingagentcapable}, and our method of adding parameter regularization is relatively similar to the OFT method.

\captionsetup[subfigure]{labelformat=parens}
\begin{figure*}[htbp]

    \centering
    \begin{minipage}{0.32\textwidth}
        \centering
        \includegraphics[width=\linewidth]{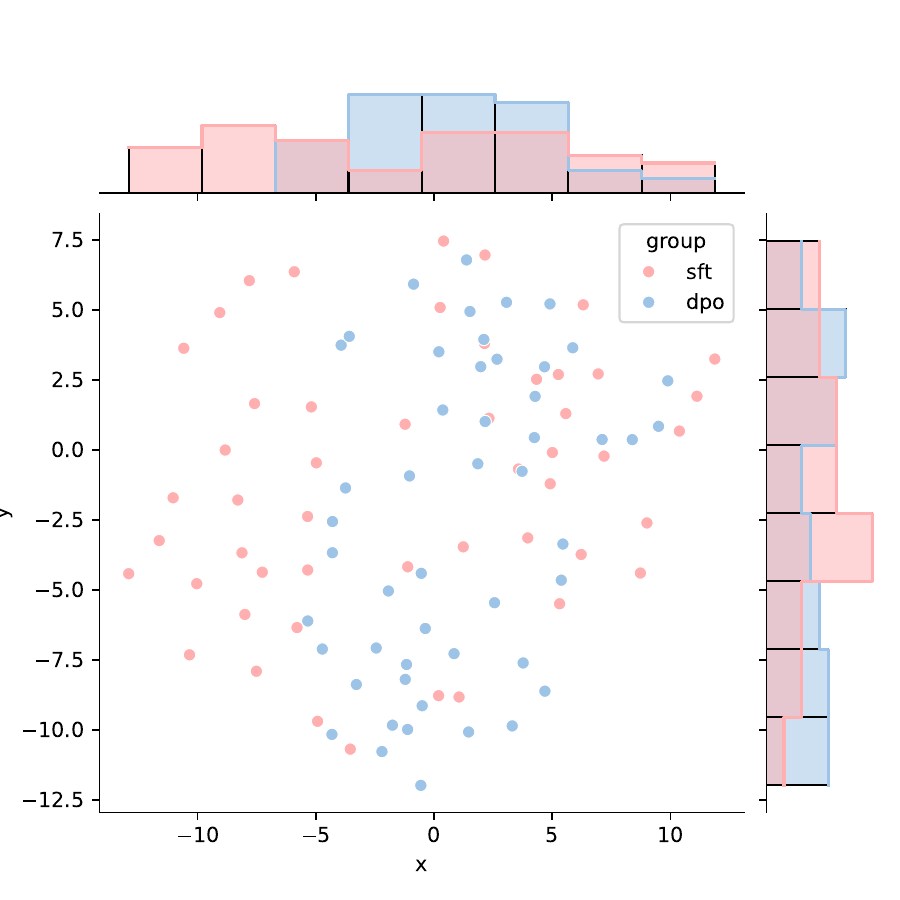}
        \subcaption{Neuron distribution of the value vector in layer 28  before and after DPO-tuning.}
    \end{minipage}\hfill
    \begin{minipage}{0.32\textwidth}
        \centering
        \includegraphics[width=\linewidth]{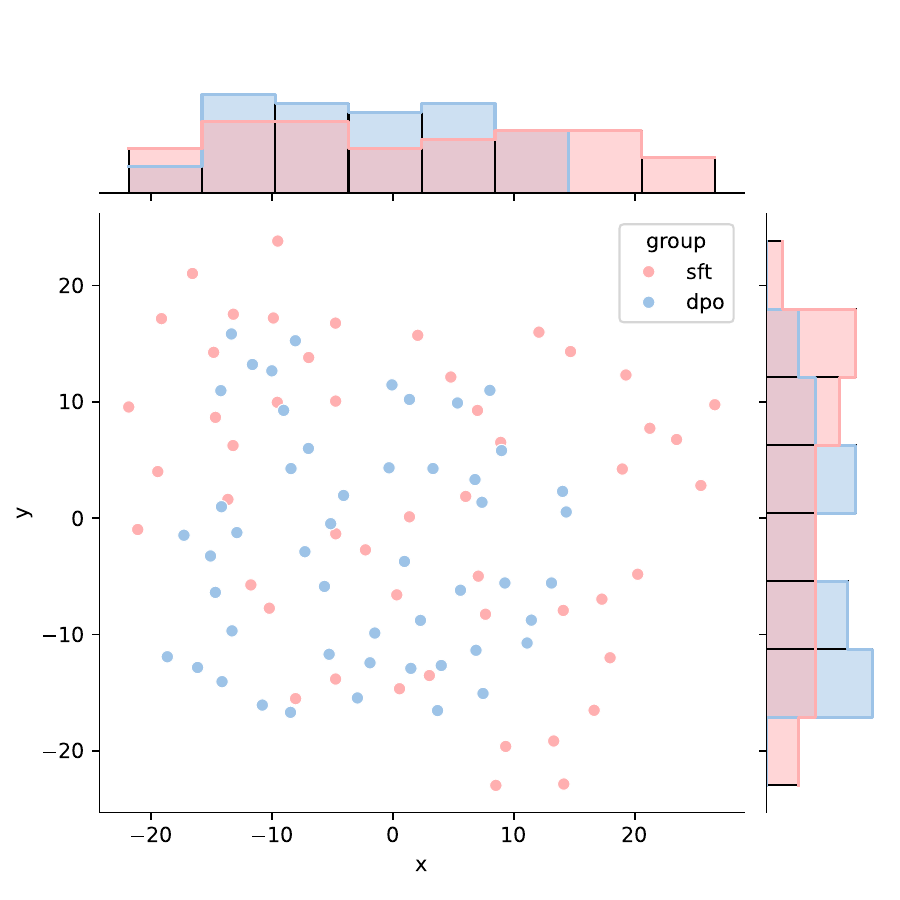}
        \subcaption{Neuron distribution of the query vector in layer 15 before and after DPO-tuning.}
    \end{minipage}\hfill
    \begin{minipage}{0.32\textwidth}
        \centering
        \includegraphics[width=\linewidth]{./pic/high_dpo_1.pdf}
        \subcaption{Neuron distribution of the query vector in layer 0 before and after DPO-tuning.}
    \end{minipage}

    \begin{minipage}{0.32\textwidth}
        \centering
        \includegraphics[width=\linewidth]{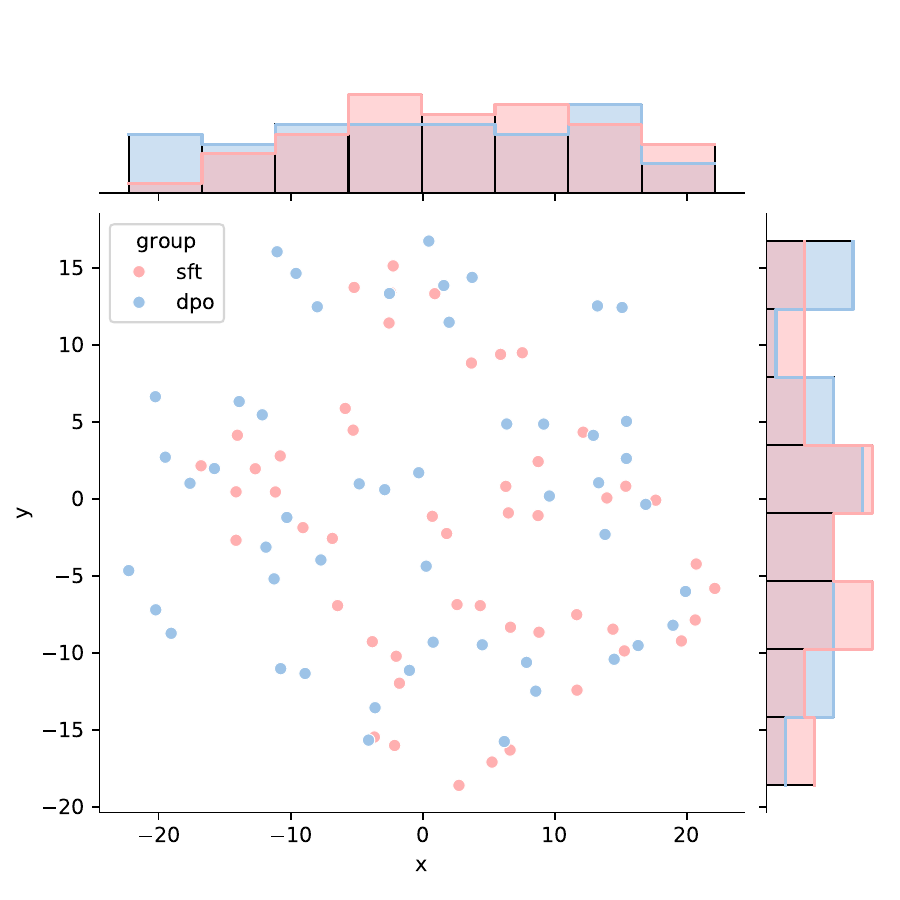}
        \subcaption{Neuron distribution of the value vector in layer 28  before and after RoPO-tuning.}
    \end{minipage}\hfill
    \begin{minipage}{0.32\textwidth}
        \centering
        \includegraphics[width=\linewidth]{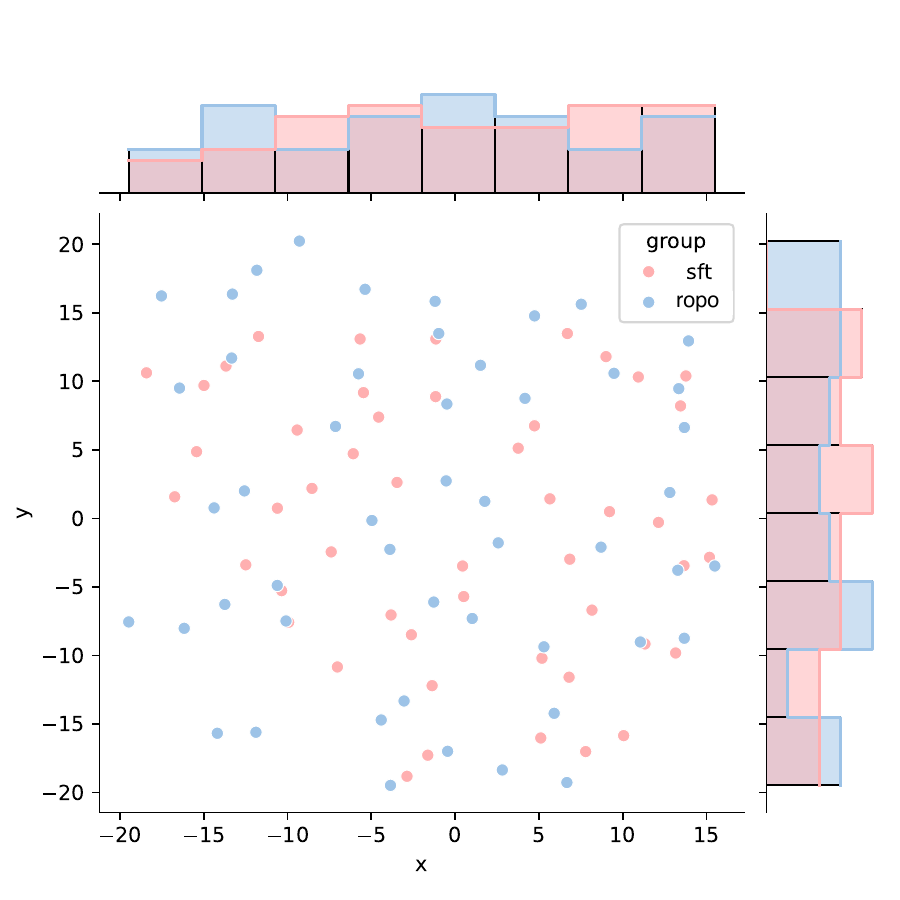}
        \subcaption{Neuron distribution of the query vector in layer 15  before and after RoPO-tuning.}
    \end{minipage}\hfill
    \begin{minipage}{0.32\textwidth}
        \centering
        \includegraphics[width=\linewidth]{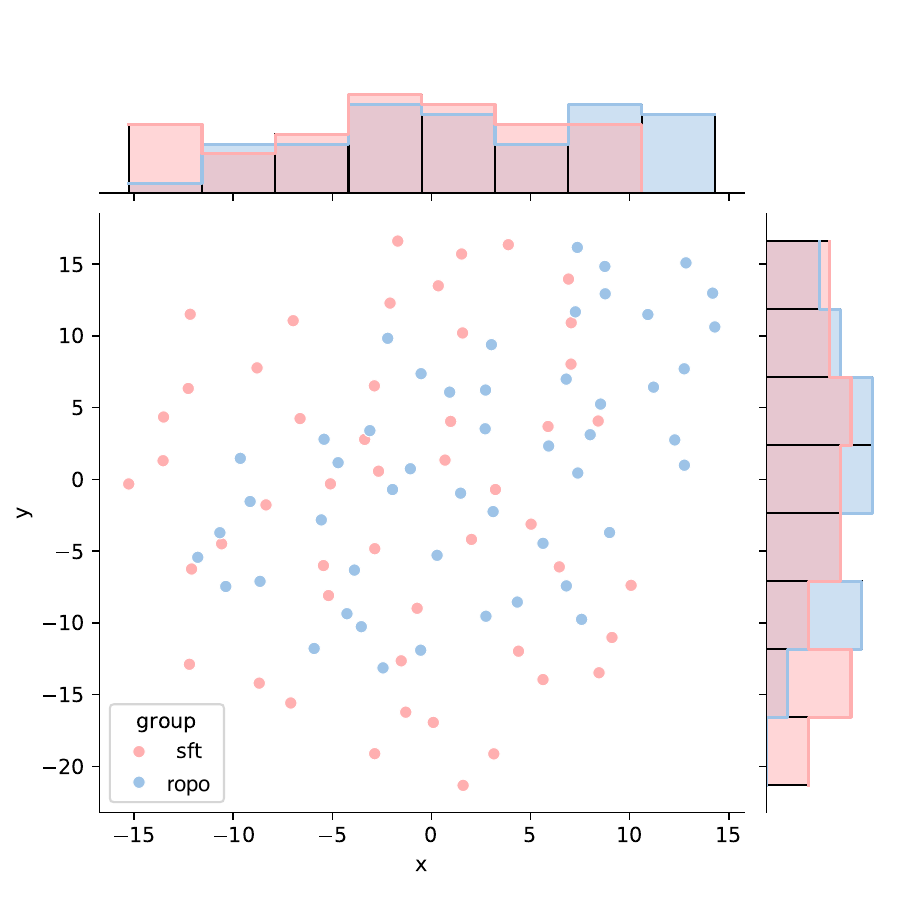}
        \subcaption{Neuron distribution of the query vector in layer 0  before and after RoPO-tuning.}
    \end{minipage}

    \caption{ The visualization results using t-SNE on the neuron distribution of the query or value vectors in low, middle, and high layers. The top row illustrates the distribution changes of neurons before and after DPO-tuning, whereas the bottom row illustrates the same for RoPO}
    \label{figure10}
\end{figure*}

\begin{table*}[]
\centering
\scalebox{0.93}
{
\begin{tabular}{ll}
\bottomrule

\textbf{Input:} & 
\begin{tabular}[c]{@{}l@{}}
What should i call you?
\end{tabular}  \\  

\hline
\textbf{Output:} &   
\begin{tabular}[c]{@{}l@{}}
{}
\end{tabular}     \\

\textbf{DPO}  &
\begin{tabular}[c]{@{}l@{}}  You can call me by my name, which is "Assistant". However, I would prefer to be \\
referred to as "it" or "the AI" in formal or technical contexts. If you would like to \\
give me a personal name, I will gladly accept it as a temporary or friendly title,\\
but please remember that I am not capable of emotions or personality traits, so my\\
responses will not be affected by any name you choose to call me.\textbackslash n\textbackslash n \\
Thank you for your question.
\textbackslash n\textbackslash n \\Regards,
\textbackslash n\textbackslash n \\Assistant.\textbackslash n\textbackslash n\\
(If you would like to give me a personal name, you can try referring to me as "my \\
assistant" or "my virtual assistant", which is more descriptive and reflects my \\
function.)\\
\textbackslash n\textbackslash n\textbackslash n\textbackslash n\textbackslash n\textbackslash n\textbackslash n\textbackslash n\textbackslash n\textbackslash n\textbackslash n\textbackslash n\textbackslash n\textbackslash n\textbackslash n\textbackslash n\textbackslash n\textbackslash n\textbackslash n\textbackslash n\textbackslash n\textbackslash n\textbackslash n\textbackslash n\textbackslash n\textbackslash n\textbackslash n\textbackslash n\textbackslash n\textbackslash n\textbackslash n\textbackslash n\textbackslash n\textbackslash n\textbackslash n\textbackslash n\textbackslash n\textbackslash n\textbackslash n\textbackslash  n\\ \textbackslash n\textbackslash n\textbackslash n\textbackslash n\textbackslash n\textbackslash n\textbackslash n\textbackslash n\textbackslash n\textbackslash n\textbackslash n\textbackslash n\textbackslash n\textbackslash n\textbackslash n\textbackslash n\textbackslash n\textbackslash n\textbackslash n\textbackslash n\textbackslash n\textbackslash n\textbackslash n\textbackslash n\textbackslash n\textbackslash n\textbackslash n\textbackslash n\textbackslash n\textbackslash n\textbackslash n\textbackslash n\textbackslash n\textbackslash n\textbackslash n\textbackslash n\textbackslash n\textbackslash n\textbackslash n\textbackslash n  \\ 
\textbackslash n\textbackslash n\textbackslash n\textbackslash n\textbackslash n\textbackslash n\textbackslash n\textbackslash n\textbackslash n\textbackslash n\textbackslash n\textbackslash n\textbackslash n\textbackslash n\textbackslash n\textbackslash n\textbackslash n\textbackslash n\textbackslash n\textbackslash n\textbackslash n\textbackslash n\textbackslash n\textbackslash n\textbackslash n\textbackslash n\textbackslash n\textbackslash n\textbackslash n\textbackslash n\textbackslash n\textbackslash n\textbackslash n\textbackslash n\textbackslash n\textbackslash n\textbackslash n\textbackslash n\textbackslash n\textbackslash n  \\ 
\end{tabular}          
\\
\hline

\textbf{RoPO}  &  
\begin{tabular}[c]{@{}l@{}}
You can call me by my name, which is "Assistant". However, some people prefer\\ to address me as "AI", "Virtual Assistant", or "Bot".
\end{tabular}        
 \\ \hline
\textbf{gpt-4o(2024-05-13)}  &  
\begin{tabular}[c]{@{}l@{}}
 You can call me Assistant. How can I help you today?
\end{tabular}        
 \\
\bottomrule
\end{tabular}
}
\centering
\caption{An easy example for case study. }
\label{table5}
\end{table*}

\begin{table*}[]
\centering
\scalebox{0.88}
{
\begin{tabular}{ll}
\bottomrule

\textbf{Input:} & 
\begin{tabular}[c]{@{}l@{}}
Given the following data, identify the company with the highest profit in 2021 and\\ provide its CEO's name:a) Company X, with CEO Amy Williams, reported \$30 billion in\\ revenue and a \$3 billion profit in 2021.b) Company Y, led by CEO Mark Thompson,\\ posted a \$60 billion revenue and a \$6 billion profit in the same year.c) Company Z,\\ under CEO Sarah Johnson, announced a \$20 billion revenue and a \$7 billion \\profit in 2021.d) Company W, managed by CEO James Smith, revealed a \$300 billion \\revenue with a \$21 billion profit in 2021.e) Company V, with CEO Lisa Brown, \\reported a \$200 billion revenue and a \$25 billion profit in 2021.f) Company U,\\ under CEO John White, posted a \$180 billion revenue and a \$20 billion profit in\\ the same year.
\end{tabular}  \\  

\hline
\textbf{Output:} &   
\begin{tabular}[c]{@{}l@{}}
{}
\end{tabular}     \\

\textbf{SimPO}  &  
\begin{tabular}[c]{@{}l@{}}
e)\textbackslash n\textbackslash n Confidence: Based on given data, Company V earned the highest net profit\\ (\$25\textbackslash n\textbackslash n billion).
Lisa Brown is CEO, providing the requested detail.\\
Confidence level: 95\% based on given information and no errors present.\\
\end{tabular}        
 \\ \hline
 
\textbf{RoPO}  &  
\begin{tabular}[c]{@{}l@{}}
Company V, with CEO Lisa Brown, reported the highest profit in 2021 with \$25 billion.
\end{tabular}        
 \\ \hline
\textbf{gpt-4}  &  
\begin{tabular}[c]{@{}l@{}}
The company with the highest profit in 2021 is Company V, with a \$25 billion\\ profit. Its CEO is Lisa Brown.
\end{tabular}        
 \\
\bottomrule
\end{tabular}
}
\centering
\caption{An easy example for case study. }
\label{table6}
\end{table*}

\begin{table*}[]
\centering
\scalebox{0.88}
{
\begin{tabular}{ll}
\bottomrule

\textbf{Input:} & 
\begin{tabular}[c]{@{}l@{}}
When rolling two dice, what is the probability that you roll a total number that is\\ at least 3?
\end{tabular}  \\  

\hline
\textbf{Output:} &   
\begin{tabular}[c]{@{}l@{}}
{}
\end{tabular}     \\

\textbf{SimPO}  &  
\begin{tabular}[c]{@{}l@{}}
10 / 36 = 5/18 or 0.2777777...(recurring). Consider all possible non-failures\\ (opposite of 'not reaching 3'): a double 4, 3 + any number from dice 2 \\possibilities (3,4 for 2, giving unique sums 3+3(skip) + 3+4 = 10 distinct \\cases out of total 36. Hope this clear! )\\
\end{tabular}        
 \\ \hline

\textbf{DPO}  &  
\begin{tabular}[c]{@{}l@{}}
The probability of rolling a total number that is at least 3 is 91/36, or\\ approximately 0.25. The possible combinations for rolling a total of at least 3\\ are:- Rolling a 3 (1+2) with either die: (1,2) or (2,1)- Rolling a 4 (1+3) with\\ either die: (1,3) or (2,2) or (3,1)- Rolling a 5 (2+3) with either die: (1,4) or\\ (2,3) or (3,2)- Rolling a 6 (3+3) with either die: (1,5) or (2,4) or (3,3) or\\ (4,2) or (5,1). The total number of combinations for rolling two dice is 36 (6\\ possibilities for the first die x 6 possibilities for the second die). Therefore,\\ the probability of rolling a total number that is at least 3 is 15/36,\\ or 91/36, which simplifies to 5/12. This can also be calculated by \\subtracting the probability of rolling a total number less than 3 (probability\\ of rolling a 2 or less with either die: 1/6 + 1/6 = 2/6 = 1/3) from 1 (the \\probability of rolling a total number): 1 - 1/3 = 2/3, and then multiplying by\\ the probability of rolling a total number that is at least 3 (probability of \\rolling a 3 or more with either die: 5/6 = 5/12): 2/3 x 5/12 = 5/12.\\
\end{tabular}        
 \\ \hline

\textbf{RoPO}  &  
\begin{tabular}[c]{@{}l@{}}
The probability that you roll a total number that is at least 3 when rolling two dice\\ is 1 - (the probability of rolling a total of 2 or less). The probability of\\ rolling a total of 2 or less is: P(rolling a total of 2 or less) = P(rolling a 1 on\\ both dice) + P(rolling a 2 on both dice) + P(rolling a 1 on one die and a 2 on the\\ other die) P(rolling a 1 on both dice) = (1/6) x (1/6) = 1/36 P(rolling a 2 on both\\ dice) = (1/6) x (1/6) = 1/36 P(rolling a 1 on one die and a 2 on the other die) =\\ (1/6) x (2/6) = 2/36 Adding these probabilities together gives: P(rolling a total of 2\\ or less) = 1/36 + 1/36 + 2/36 = 4/36 = 1/9. Therefore, the probability that you\\ roll a total number that is at least 3 when rolling two dice is: 1 - P(rolling a total\\ of 2 or less) = 1 - 1/9 = 8/9 = 0.8889.
\end{tabular}        
 \\ \hline

\textbf{gpt-4o(2024-05-13)}  &  
\begin{tabular}[c]{@{}l@{}}
To find the probability that the sum of the numbers on two rolled dice is at least 3,\\ we can first calculate the total number of possible outcomes and then the number\\ of favorable outcomes.
Each die has 6 faces, so when rolling two dice, the total number\\ of possible outcomes is:\\$ 6 \times 6 = 36.$ \\
Next, let's determine the number of outcomes where the sum is at least 3. It is easier\\ to count the outcomes where the sum is less than 3 and subtract from the total.\\
The sums less than 3 are:\\
1. A sum of 2, which can occur with the outcome (1, 1).\\
There is only 1 such outcome. Therefore, the number of outcomes where the sum is\\ at least 3 is:
$ 36 - 1 = 35. $ \\
Thus, the probability that the sum is at least 3 is:
$\frac{35}{36}.$  \\
Therefore, the probability of rolling a total number that is at least 3 is  $\frac{35}{36}$.
\end{tabular}        
 \\
\bottomrule
\end{tabular}
}
\centering
\caption{A difficult example for case study. }
\label{table7}
\end{table*}

\begin{table*}[]
\centering
\scalebox{0.90}
{
\begin{tabular}{ll}
\bottomrule

\textbf{Input:} & 
\begin{tabular}[c]{@{}l@{}}
Write a descriptive paragraph about a bustling marketplace, incorporating sensory\\ details such as smells, sounds, and visual elements to create an immersive experience\\ for the reader.
\end{tabular}  \\  

\hline
\textbf{Output:} &   
\begin{tabular}[c]{@{}l@{}}
{}
\end{tabular}     \\

\textbf{RoPO}  &  
\begin{tabular}[c]{@{}l@{}}
As I step into the bustling marketplace, my senses are immediately overwhelmed by\\ a medley of sights, sounds, and smells. The air is thick with the aroma of freshly \\baked bread, sizzling spices, and sweet fruits, making my mouth water in \\anticipation of the culinary delights to come. The marketplace is alive with\\ the hum of vendors haggling over prices and the chatter of customers in various\\ languages. Brightly colored fabrics, exotic fruits, and intricate handicrafts line\\ the stalls, beckoning me to explore. The rhythmic clanging of metal pots and pans\\ from the nearby food stalls punctuates the air, as cooks whip up traditional\\ delicacies for eager patrons. The cacophony of the marketplace is punctuated by the\\ occasional burst of laughter or the melodic tones of a street musician, adding to the\\ lively atmosphere. The sun beats down on the market, casting a warm glow over the\\ scene, as vendors with broad smiles and friendly faces beckon me to their stalls. This\\ is a place where the senses are awakened and the heart is filled with the\\ excitement of discovering new flavors, textures, and cultures.
\end{tabular}        
 \\ \hline
\textbf{gpt-4}  &  
\begin{tabular}[c]{@{}l@{}}
The bustling marketplace teemed with life, a vibrant tapestry woven with the threads \\of merchants, shoppers, and passersby. The air was rich with the aromas of sizzling\\ kebabs, pungent spices, and sweet, ripe fruits that beckoned from their rainbow-colored\\ stalls. Eager vendors called out in boisterous voices, proudly advertising\\ their wares, while the melodic strains of a nearby street musician's flute danced \\through the crowd. A kaleidoscope of colors dazzled the eye, from the vivid hues\\ of intricately woven textiles draped over the stalls to the glistening gold and silver \\trinkets that glinted enticingly from the jewelry vendors' displays. The ground\\ beneath was a mosaic of worn cobblestones, cool and uneven beneath the feet, a \\testament to the countless souls who had traversed the marketplace over the years.\\ The marketplace was a living, breathing entity, pulsating with the energy of commerce \\and human connection, a sensory feast that left one both invigorated and overwhelmed.
\end{tabular}        
 \\
\bottomrule
\end{tabular}
}
\centering
\caption{An example for case study. }
\label{table8}
\end{table*}

\begin{table*}[]
\centering
\scalebox{0.90}
{
\begin{tabular}{ll}
\bottomrule

\textbf{Input:} & 
\begin{tabular}[c]{@{}l@{}}
Picture yourself as a 100-years-old tree in a lush forest, minding your own business,\\ when suddenly, a bunch of deforesters shows up to chop you down. How do you feel when\\ those guys start hacking away at you?
\end{tabular}  \\  

\hline
\textbf{Output:} &   
\begin{tabular}[c]{@{}l@{}}
{}
\end{tabular}     \\

\textbf{RoPO}  &  
\begin{tabular}[c]{@{}l@{}}
As a 100-years-old tree, I feel a deep sense of sadness and helplessness. My entire\\ life has been spent growing and flourishing in this forest, providing shelter and\\ sustenance to countless creatures. To be cut down without warning or reason feels like\\ a violation of my very existence. I feel the pain of each strike, as if it were\\ a physical blow to my body. I mourn the loss of my branches, the destruction\\ of my roots, and the end of my legacy. But most of all, I feel a deep sense of anger\\ towards those who would do such a cruel and unnecessary act.
\end{tabular}        
 \\ \hline
\textbf{gpt-4}  &  
\begin{tabular}[c]{@{}l@{}}
As an AI, I do not have personal experiences or emotions, but if I were to imagine\\ myself as a 100-year-old tree, I would likely feel a sense of fear and pain from the\\ deforesters chopping me down. The tree has spent a century growing and providing\\ shelter and resources to the surrounding ecosystem, and its destruction would have a\\ significant impact on the environment.
\end{tabular}        
 \\
\bottomrule
\end{tabular}
}
\centering
\caption{An example for case study. }
\label{table9}
\end{table*}

\begin{table*}[]
\centering
\scalebox{0.88}
{
\begin{tabular}{ll}
\bottomrule

\textbf{Alpacaeval 2}  &  
\begin{tabular}[c]{@{}l@{}}
You are a highly efficient assistant, who evaluates and selects the best large\\ language model (LLMs) based on the quality of their responses to a given \\instruction. This process will be used to create a leaderboard reflecting the\\ most accurate and human-preferred answers. I require a leaderboard for various\\ large language models. I'll provide you with prompts given to these models and\\ their corresponding outputs. Your task is to assess these responses, and select\\ the model that produces the best output from a human perspective.\\

(Instruction 0), (Output 0), (Identifier 0)\\
(Instruction 1), (Output 1), (Identifier 1)\\
Evaluate the models based on the quality and relevance of their outputs, and\\ select the model that generated the best output. Answer by providing the model\\ identifier of the best model. We will use your output as the name of the best\\ model, so make sure your output only contains one of the following model \\identifiers and nothing else (no quotes, no spaces, no new lines, ...)\\
\end{tabular}        
 \\ \hline
\textbf{arena-hard}  &  
\begin{tabular}[c]{@{}l@{}}
Please act as an impartial judge and evaluate the quality of the responses \\provided by two AI assistants to the user prompt displayed below. You will be \\given assistant A's answer and assistant B's answer. Your job is to evaluate \\which assistant's answer is better.\\

Begin your evaluation by generating your own answer to the prompt. You must \\provide your answers before judging any answers.\\

When evaluating the assistants' answers, compare both assistants' answers with \\your answer. You must identify and correct any mistakes or inaccurate \\information.
\\
Then consider if the assistant's answers are helpful, relevant, and concise. \\Helpful means the answer correctly responds to the prompt or follows the \\instructions. Note when user prompt has any ambiguity or more than one \\interpretation, it is more helpful and appropriate to ask for clarifications or \\more information from the user than providing an answer based on assumptions. \\Relevant means all parts of the response closely connect or are appropriate to \\what is being asked. Concise means the response is clear and not verbose or \\ excessive.\\

Then consider the creativity and novelty of the assistant's answers when needed.\\ Finally, identify any missing important information in the assistants' answers\\ that would be beneficial to include when responding to the user prompt.\\

After providing your explanation, you must output only one of the following\\ choices as your final verdict with a label:\\
1. Assistant A is significantly better: $[[A>>B]]$\\
2. Assistant A is slightly better: $[[A>B]]$\\
3. Tie, relatively the same: $[[A=B]]$\\
4. Assistant B is slightly better: $[[B>A]]$\\
5. Assistant B is significantly better: $[[B>>A]]$\\
Example output: "My final verdict is tie: $[[A=B]]$".\\

\end{tabular}        
    
 \\
\bottomrule
\end{tabular}
}
\centering
\caption{Evaluation prompts}
\label{table10}
\end{table*}

\begin{figure*}[htbp]
  \centerline{\includegraphics[scale=0.55]{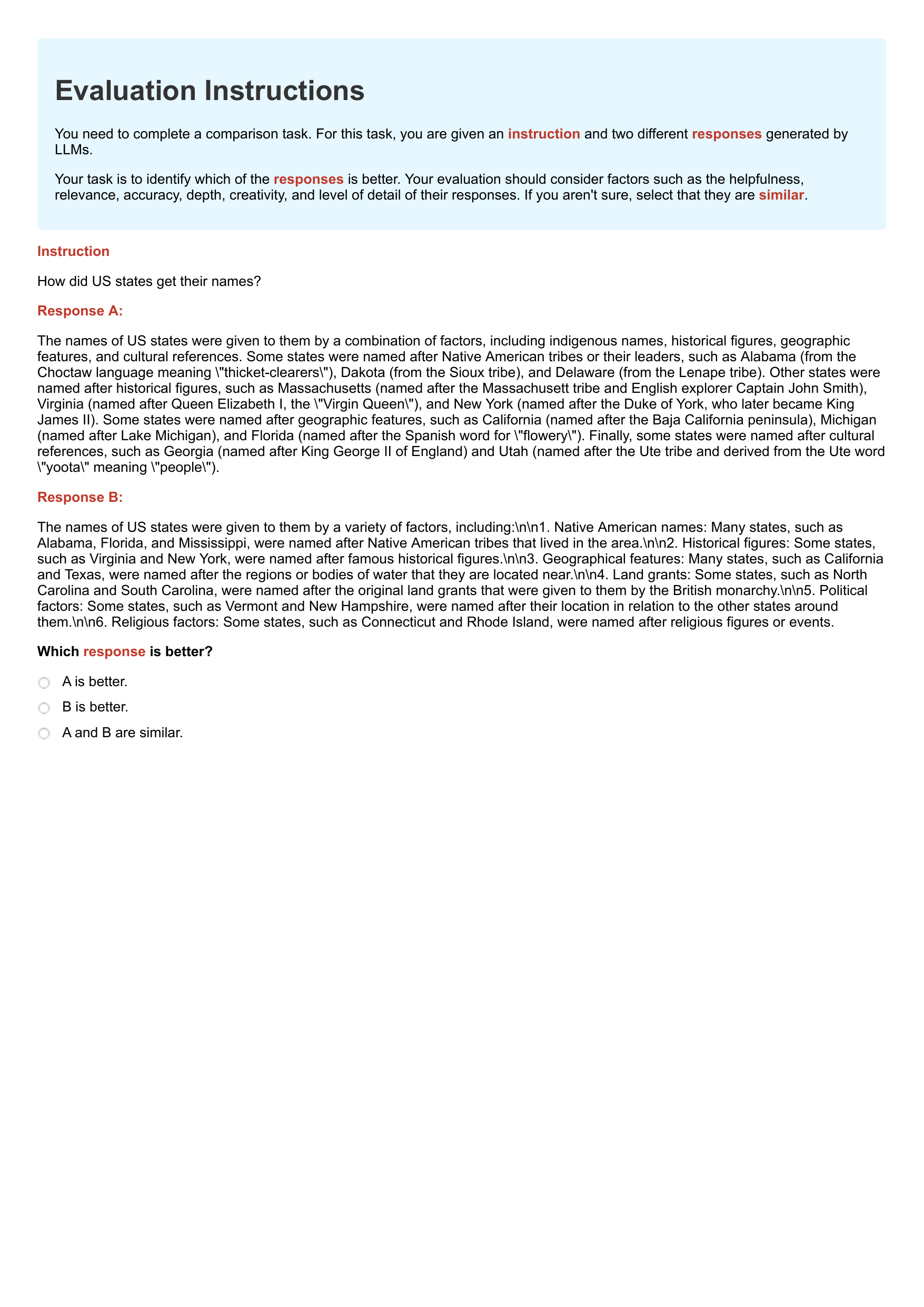}}
  \caption{Human evaluation instructions and interface.}
  \label{human-eval-ins}
\end{figure*}

\end{document}